%% file: iclr2027_conference.tex
\documentclass{article} 
\PassOptionsToPackage{dvipsnames}{xcolor}
\usepackage{iclr2027_conference,times}
\usepackage{makecell}
\iclrfinalcopy

\input{math_commands.tex}

\usepackage{hyperref}
\usepackage{url}
\usepackage{stfloats}

\usepackage{enumitem}
\usepackage{booktabs}
\usepackage{amsfonts}
\usepackage{amsmath}
\usepackage{amssymb}
\usepackage{multirow}
\usepackage{subcaption}
\usepackage{graphicx}
\usepackage{tabularx}
\usepackage{threeparttable}
\graphicspath{{cc_Figures/}{Figures/}}
\usepackage{algorithm}
\usepackage[noend]{algorithmic}
\usepackage[utf8]{inputenc}
\usepackage[most]{tcolorbox}
\usepackage{xcolor}
\usepackage{colortbl}
\definecolor{darkgray}{RGB}{60, 60, 60}   

\definecolor{TopGray}{HTML}{EFEFEF}   
\definecolor{NRBlue}{HTML}{D9E2EC}   
\definecolor{RGreen}{HTML}{F4ECF7}   

\definecolor{SysHighlight}{HTML}{F4ECF7} 
\definecolor{BaseGray}{HTML}{EFEFEF}    

\definecolor{SectionBg}{HTML}{D9E2EC} 
\definecolor{SubBg}{HTML}{F4ECF7}     

\newtcolorbox{stepbox}{
    colback=gray!5,      
    colframe=black,      
    boxrule=0.7pt,
    sharp corners,       
    left=5pt, right=5pt, top=5pt, bottom=5pt,
    fontupper=\small\ttfamily 
}

\newtcolorbox{promptbox}{
    enhanced,
    breakable,
    colback=gray!5,
    colframe=black,
    boxrule=0.7pt,
    sharp corners,
    left=5pt, right=5pt, top=5pt, bottom=5pt,
    fontupper=\scriptsize\ttfamily,
    before skip=4pt,
    after skip=4pt
}

\newtcolorbox{domainbox}{
    colback=gray!3,
    colframe=darkgray,
    boxrule=0.5pt,
    sharp corners,
    left=5pt, right=5pt, top=5pt, bottom=5pt,
    fontupper=\small
}

\usepackage{listings}
\lstdefinestyle{casecode}{
  basicstyle=\scriptsize\ttfamily,
  keywordstyle=\color{RoyalBlue!80!black}\bfseries,
  commentstyle=\color{gray!70!black}\itshape,
  language=Python,
  columns=fullflexible, keepspaces=true,
  breaklines=true, breakatwhitespace=true,
  frame=none, aboveskip=2pt, belowskip=2pt,
  escapeinside={(*@}{@*)}
}

\usepackage{tikz}
\usetikzlibrary{calc, tikzmark}
\usetikzlibrary{arrows.meta, positioning, shapes.geometric}
\usetikzlibrary{decorations.pathreplacing}
\usepackage{arydshln}

\usepackage[dvipsnames]{xcolor} 
\usepackage{algorithm}

\usepackage{tcolorbox}          
\tcbuselibrary{skins, breakable}

\usepackage{wrapfig}
\usepackage{needspace} 
\usepackage{placeins} 
\usepackage{float}    
\usepackage{booktabs} 

\usepackage{tabularx}
\usepackage{ragged2e}
\usepackage[table]{xcolor}

\newtcolorbox{InnerBox}{
    enhanced, breakable,
    colback=RoyalBlue!5,        
    colframe=RoyalBlue!80!black,
    arc=0mm,                    
    leftrule=2.5pt,             
    rightrule=0pt, toprule=0pt, bottomrule=0pt, 
    left=4pt, right=4pt, top=4pt, bottom=4pt,   
    boxsep=0pt,
    before=\par\vspace{4pt}\noindent, after=\par\vspace{4pt}
}

\newtcolorbox{OuterBox}{
    enhanced, breakable,
    colback=BurntOrange!8,      
    colframe=BurntOrange!90!black, 
    arc=0mm,
    leftrule=2.5pt,
    rightrule=0pt, toprule=0pt, bottomrule=0pt,
    left=4pt, right=4pt, top=4pt, bottom=4pt,
    boxsep=0pt,
    before=\par\vspace{4pt}\noindent, after=\par\vspace{4pt}
}

\title{MetaBench-Harness: Unlocking End-to-End Optimization of Benchmark Harnesses}

\definecolor{boxbg}{HTML}{F1F4F7}
\newcommand{\sys}{\textsc{MetaBench-Harness}}

\begin{document}

\pagestyle{plain}

\addtolength{\topmargin}{-0.28cm}
\addtolength{\textheight}{0.28cm}

\begin{tcolorbox}[
    enhanced,
    colback=boxbg,       
    colframe=boxbg,      
    arc=4mm,             
    boxrule=0pt,         
    left=20pt, right=20pt, top=20pt, bottom=20pt, 
    fonttitle=\bfseries
]

{\LARGE \bfseries MetaBench-Harness: Unlocking End-to-End \\ Optimization of Benchmark Harnesses \par}
\vspace{1.5em}

{\normalsize \bfseries
Xuanjun Chen\textsuperscript{1,2}, Hua-Hsuan Chen\textsuperscript{1,2}, Wei-Chung Lu\textsuperscript{1}, Yinghao Ma\textsuperscript{3} \\[0.5ex]
Jyh-Shing Roger Jang\textsuperscript{1}, Hung-yi Lee\textsuperscript{1,2} \par}
\vspace{1em}

{\normalsize
\textsuperscript{1} National Taiwan University \hspace{0.5em}
\textsuperscript{2} NTU AI-Core \hspace{0.5em}
\textsuperscript{3} Queen Mary University of London 
\par}
\vspace{2em}

Rapid progress in Large Language Models (LLMs) is saturating static benchmarks faster than they can be designed. 
While existing automated evolution frameworks attempt to generate harder questions by perturbing individual tasks, they remain constrained by rigid, hard-coded generation rules. 
Moving beyond the evolution of isolated tasks, we propose to optimize the benchmark generation workflow itself end to end with \textsc{MetaBench-Harness}, a dual-loop search framework.
Specifically, the inner loop utilizes a benchmark harness to generate a new benchmark in each round, while the outer meta-harness orchestration layer iteratively refines and searches over harness implementations based on historical evolution trajectories. 
By applying \textsc{MetaBench-Harness} to the competitive programming CodeContests and Olympiad mathematics AIME-2024 datasets, we demonstrate that the evolved benchmarks are challenging and discriminative for frontier models. 
Trajectory and quality analyses verify that \textsc{MetaBench-Harness} enables multi-dimensional evolution, steadily improving evolution reasonableness, benchmark competency, and evaluator robustness across successive rounds. Furthermore, case studies reveal its effective utilization of diverse difficulty levers to reframe problems and elevate required capabilities. Ultimately, this work provides a solution to the pressing challenge of benchmark saturation. 
\vspace{2em}

\noindent\textbf{Date:} Sep 27, 2026 \hfill 
\raisebox{-0.5\height}{\includegraphics[height=1.5em]{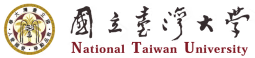}} \hspace{1em}
\raisebox{-0.5\height}{\includegraphics[height=1.5em]{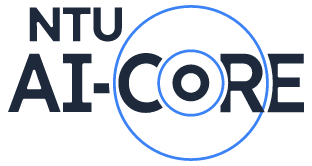}} \hspace{1em}
\raisebox{-0.5\height}{\includegraphics[height=1.5em]{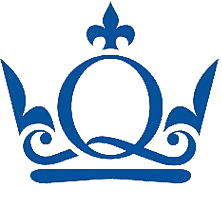}} \hspace{4.0em}
\end{tcolorbox}

\section{Introduction}
\label{sec:intro}

Recent breakthroughs in Large Language Models (LLMs) \citep{achiam2023gpt} have showcased exceptional capabilities across diverse applications, 
from creative text generation to intricate problem-solving. 
Consequently, dynamic and rigorous evaluation frameworks have emerged as a vital research domain \citep{lou2025aaar, chang2024llm_survey}. 
Effective assessment frameworks offer deep insights into model strengths and limitations, assist practitioners in selecting suitable LLMs \citep{zheng2023judging}, and provide directional guidance for future architectural and algorithmic improvements, 
as highlighted in recent taxonomies addressing the evaluation trilemma \citep{Tiwari2026RethinkingLE}. 

\begin{figure*}[t]
\centering
\includegraphics[width=0.85\columnwidth]{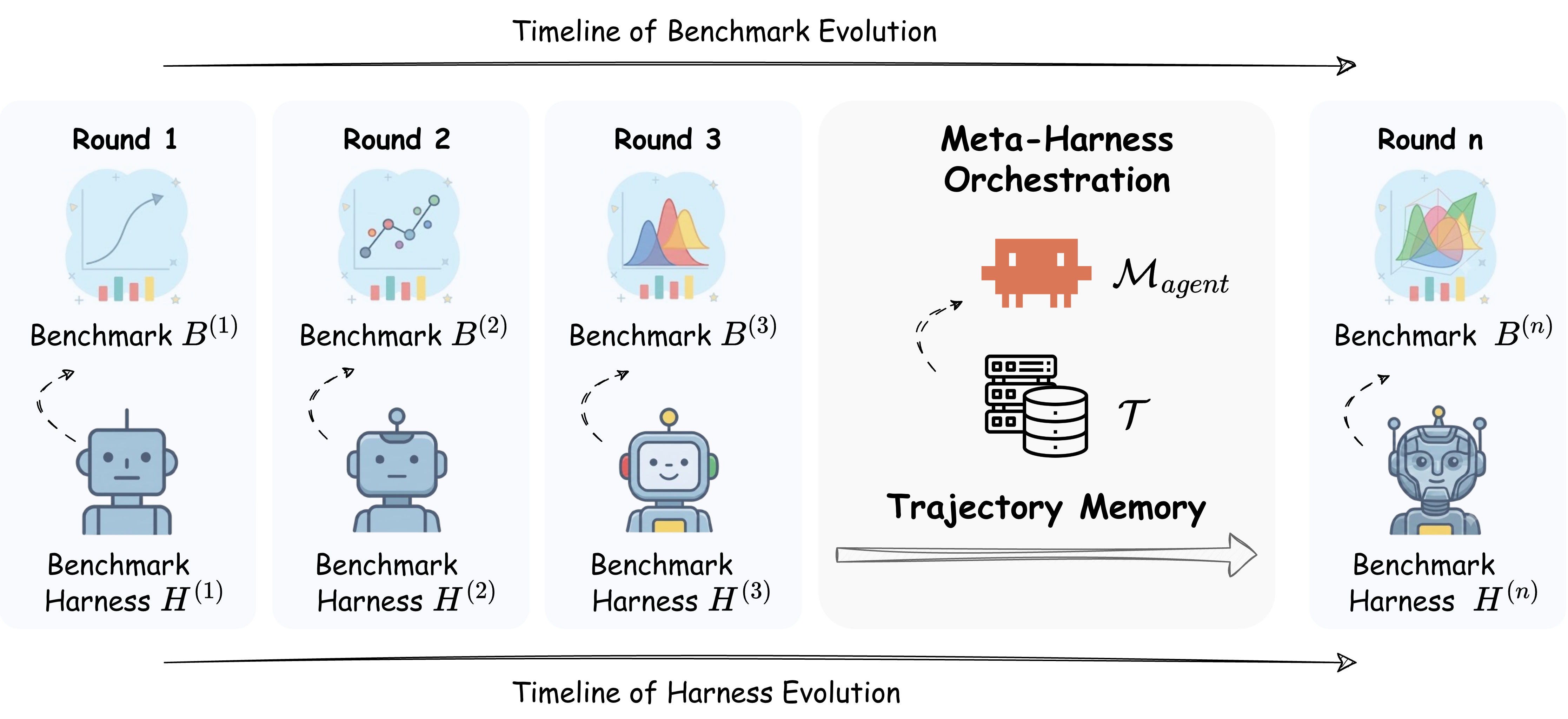}
\caption{Overview of the \sys\ framework.}
\label{fig:overview}
\end{figure*}
While various benchmarks have been introduced to evaluate LLMs, traditional static datasets are increasingly inadequate due to the rapid evolution of model capabilities and widespread data contamination \citep{cheng2025survey}, which occurs when public test sets inadvertently infiltrate training corpora and result in skewed evaluations. Although existing methods attempt to address these challenges through perplexity-based metrics or fine-grained semantic detection \citep{di2026tracer}, they remain limited by an over-reliance on static rules or poor task generalization. 
Consequently, while dynamic paradigms have emerged \citep{shi2026judgeagent}, preventing benchmark saturation through continuous and robust evolution remains a critical and largely unresolved challenge.

To meet this need, recent efforts have increasingly turned toward automated and self-evolving benchmarks designed to renew evaluation. However, current approaches still exhibit several limitations: frameworks like Benchmark Self-Evolving \citep{wang-etal-2025-benchmark} and AutoEvoEval \citep{wu2025autoevoeval} apply predefined atomic operations that primarily probe robustness rather than genuinely scaling task complexity; specialized solutions like AdamMeme \citep{chen2025adammeme} are restricted to narrow domains such as meme harmfulness; EvoCodeBench \citep{li2024evocodebench} relies on fixed ingestion schedules rather than model-driven exploration; and TRACE \citep{guo2026towards_trace} is confined to test-time exploration within specific real environments. 
However, reliance on these rigid and domain-specific mechanisms fails to provide a comprehensive solution to the ongoing risk of benchmark saturation. 

Overcoming these fragmented limitations requires moving beyond ad-hoc solutions. 
A practical benchmark generation framework must be grounded in three core pillars that directly address the aforementioned gaps: 
(1) \textbf{Evolution}, replacing rigid fixed processes and environment-restricted limits with dynamic, model-driven controllers that continuously adapt as target LLMs advance;
(2) \textbf{Discriminability}, ensuring generated queries effectively test underlying capabilities rather than merely increasing difficulty, moving beyond superficial robustness checks; and
(3) \textbf{Diversity}, ensuring that the evolved benchmarks and problems encompass diverse problem-solving techniques, target varying capability dimensions across tasks, and dynamically utilize different difficulty levers, rather than relying on uniform or narrow modifications. 
Our main contributions are threefold:
\begin{itemize}
    \item We propose \textsc{MetaBench-Harness} (Figure \ref{fig:overview}), a framework that operates over dual evolving timelines of benchmarks and harnesses. It orchestrates multi-round harness generation by feeding accumulated trajectory memory back into the search process, dynamically controlling access to prior code and execution traces to improve LLM benchmark quality. 

    \item Applying \textsc{MetaBench-Harness} to the CodeContests and AIME-2024 benchmarks demonstrates its superior effectiveness compared to previous static evolution methods in benchmark generation (Table~\ref{tab:evolution_method}), while consistently producing challenging and discriminative tasks that induce performance degradation in prominent LLM systems (Table~\ref{tab:consolidated_results}, ~\ref{tab:token_length_both_datasets}). 
 
    \item Through comprehensive analyses across trajectories, quality assessment, and quantitative case studies, we reveal that our framework not only steadily improves benchmark competency and evaluator robustness, but also effectively utilizes difficulty levers to reframe problems. Ultimately, it provides a solution to benchmark saturation.
\end{itemize}

\section{Related Work}
\label{sec:related_work}

\textbf{Static vs. Dynamic Benchmark Generation. }
Recent developments in dynamic evaluation seek to counter the swift saturation of static datasets. Early efforts like Benchmark Self-Evolving \citep{wang-etal-2025-benchmark} and AutoEvoEval \citep{wu2025autoevoeval} use predefined atomic operations for NLP reasoning and multiple-choice QA, testing surface-level robustness rather than scaling task complexity. Specialized frameworks target specific niches, such as AdamMeme \citep{chen2025adammeme} leveraging multi-agent collaboration for safety probing, or EvoCodeBench \citep{li2024evocodebench} refreshing software benchmarks at rigid temporal intervals without active model feedback.
While advanced test-time exploration approaches like TRACE \citep{guo2026towards_trace} pioneer autonomous task generation, they remain bound by predefined generative pathways. Overcoming these restrictions requires a fully dynamic pipeline. Through an outer-loop meta-harness orchestration that selectively controls the proposer's access to historical code and execution traces, \textsc{MetaBench-Harness} enables a truly dynamic benchmark generation pipeline that continuously adapts to target LLM advancements. 

\textbf{Adaptive Utilization of External Knowledge. }
Several studies demonstrate that models benefit from treating large repositories or lengthy contexts as dynamically accessed external assets rather than consuming them in a single pass. While early paradigms include retrieval-augmented generation \citep{lewis2020retrieval} and alternating retrieval-reasoning \citep{trivedi2023interleaving}, recent advances further introduce active associative discovery \citep{li2026codarag} for fragmented evidence and targeted delta planning \citep{chou2026only} for information gaps. Together with memory-enabled agents \citep{packer2023memgpt} and recursive models \citep{zhang2025recursive}, these methods highlight a shift toward adaptive context interaction. 
\sys\ applies this retrieval pattern within a benchmark-driven harness engineering loop, granting the proposer unrestricted access to historical code, scores, and execution traces to discover and improve end-to-end context handling procedures.

\textbf{Search Over Executable Code. }
Modern approaches increasingly explore executable code spaces to discover functions, workflows, or agent architectures. Early investigations integrated large models as mutation and crossover engines within evolutionary program search \citep{lehman2023evolution}. 
Subsequent techniques advanced to evolving specific functions inside fixed code frameworks \citep{romera2024mathematical}, deploying meta-agents to synthesize new agents from historical findings \citep{hu2025automated}, or optimizing workflow graphs for agentic systems \citep{zhang2025aflow}. 
Another research direction focuses on memory structures that enable continual-learning agents to retain information across multiple tasks \citep{zhang2026memevolve, xiong2026learning}. 
Closely related is Meta-Harness \citep{lee2026metaharness}, which searches over domain-specific task harnesses using a minimal outer loop with unrestricted filesystem access. 
While Meta-Harness focuses on solving individual downstream tasks for leaderboard performance, our work shifts the objective toward the automated discovery and evolution of benchmark generation pipelines.
By leveraging an open-ended search strategy, \sys\ enables a comprehensive exploration of full benchmark harness implementations, allowing the generation framework itself to continuously self-evolve.

\section{\textsc{MetaBench-Harness}: Harness for Benchmark Generation}
\label{sec:design_details}

This section details \textsc{MetaBench-Harness}, our proposed method for automated benchmark generation. As shown in Figure~\ref{fig:overview}, the proposed design integrates three core components into a pipeline: the benchmark harness, which acts as an executable program that dynamically transforms and constructs the target benchmark from seeds; a trajectory memory that records historical optimization trajectories; and a meta-harness orchestration that drives the iterative refinement process. 

\begin{figure*}[t]
\centering
\includegraphics[width=1\textwidth]{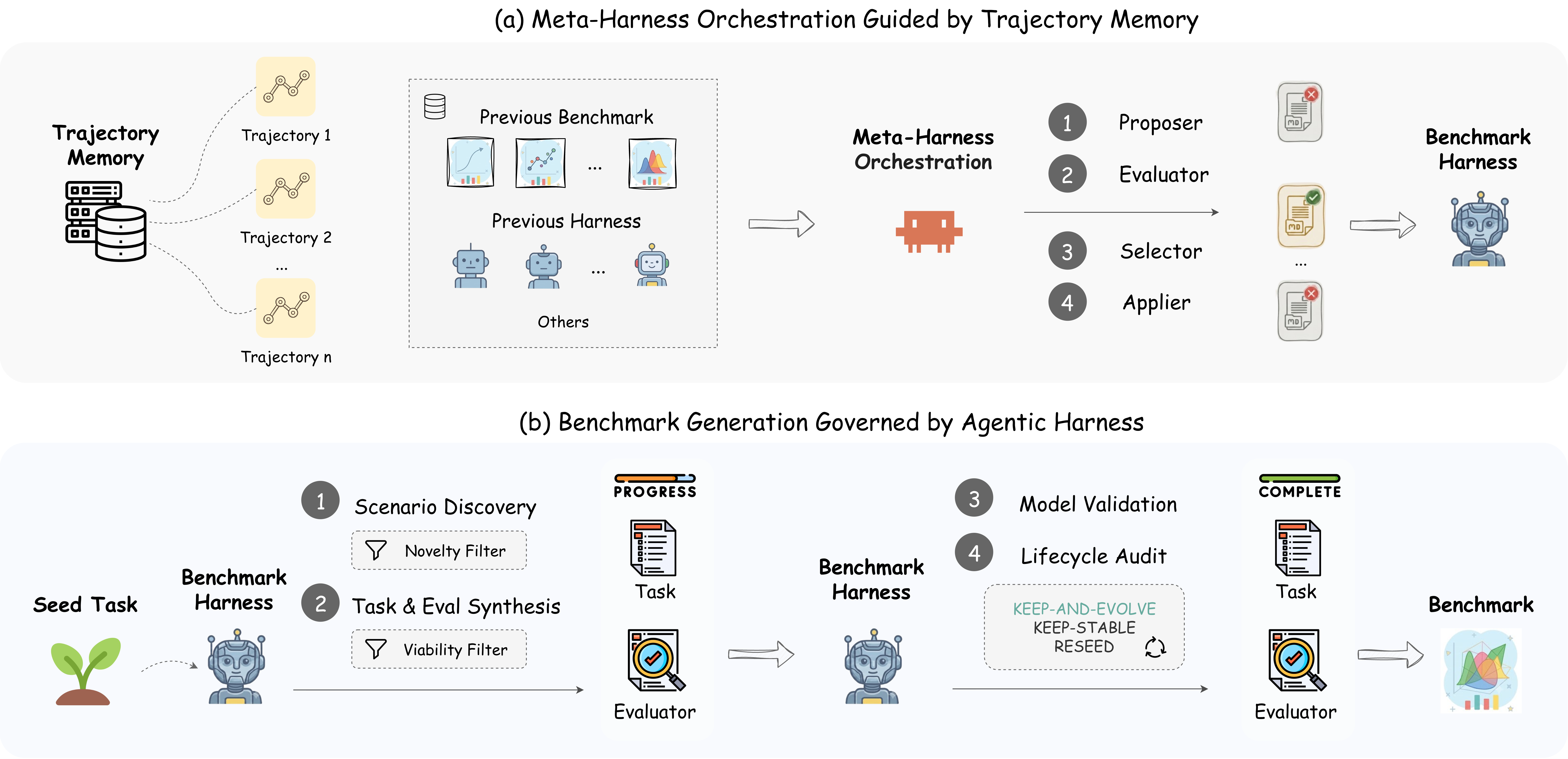}
\caption{The dual-loop procedure of \sys. 
(a) Meta-Harness Orchestration: The outer loop where the LLM agent, conditioned on trajectory memory, proposes harness updates. 
(b) Benchmark Generation: The inner loop where the evolved harness drives benchmark synthesis. }
\label{fig:pipeline}
\end{figure*}

\textbf{Objective.} Our framework employs an LLM agent $\mathcal{M}_{\text{agent}}$ to iteratively propose benchmark harnesses.  
The objective is straightforward: to find the harness that poses challenging questions to the underlying proxy models while maintaining discriminability among the models. 
To guide the optimization, we maintain a trajectory memory $\mathcal{T}$ that stores past iterations of proposed harnesses, generated benchmarks, and their feedback. At each iteration, the agent conditions on this memory to generate a new candidate harness $H \sim p_{\mathcal{M}}(\mathcal{T})$. The objective of our benchmark harness optimization is to find the optimal harness $H^*$ that maximizes the expected reward:
$$
H^* = \arg\max_{H} \mathbb{E}_{B \sim H(B_{\text{seed}})} \left[ r(s_1, s_2) \right]
$$
where $B_{\text{seed}}$ is the initial seed benchmark, and $B \sim H(B_{\text{seed}})$ denotes the target benchmark instantiated by executing $H$ as a stateful program. The variables $s_1$ and $s_2$ represent the respective validation scores of two proxy models of varying capabilities, denoted as $\mathcal{M}_1$ and $\mathcal{M}_2$, when validated on $B$. 
Here, $r(s_1, s_2)$ serves as the \textit{extrinsic execution reward}, an empirical metric that scores the generated benchmark, where it jointly maximizes the score discrepancy between models to ensure discriminability while minimizing their average performance to guarantee challenge, so that a higher $r$ signifies a more ideal benchmark harness. 
To realize this optimization objective, our framework introduces a dual-level workflow (Figure \ref{fig:pipeline}). 
The system operates through two interconnected stages: an outer loop for Meta-Harness Search Orchestration (Figure \ref{fig:pipeline}a), and an inner loop for Benchmark Generation Governed by Agentic Harness (Figure \ref{fig:pipeline}b).

\textbf{Meta-Harness Orchestration Guided by Trajectory Memory.} 
To drive the iterative evolution of the benchmark harness over time, the meta-harness orchestration mechanism employs an LLM-based agent $\mathcal{M}_{\text{agent}}$ conditioned on an accumulated trajectory memory $\mathcal{T}$ to propose candidate harness updates. 
Unlike prior systems relying on manual search heuristics to drive improvements, we delegate diagnosis and harness generation directly to the agent through a cohesive four-step workflow. 
First, acting as the \textit{proposer}, the agent inspects uncompressed raw artifacts from $\mathcal{T}$, including past harness code, generated benchmarks, and execution traces, to generate candidate updates to the incumbent harness $H^{(t-1)}$. 
Second, an \textit{evaluator} simply scores and ranks these proposals to help identify promising candidates. Third, a \textit{selector} chooses the best benchmark harness candidate from the rankings. 
Finally, the \textit{applier} applies the approved change to instantiate the evolved benchmark harness $H^{(t)}$, which subsequently drives the inner execution loop to generate the new benchmark checkpoint $B^{(t)}$. 
Relying on the uncompressed, accumulated trajectory memory, our meta-harness orchestration enables precise error diagnosis, prevents redundant search paths, and effectively guides the selection toward optimal benchmark harness evolution across generations. 

\textbf{Benchmark Generation Governed by Agentic Harness.} 
To generate high-quality benchmarks using the evolving benchmark harness, the inner execution loop runs once per optimization round $t$ using the updated harness $H^{(t)}$, a seed benchmark $B_{\text{seed}}$, and proxy models $\mathcal{M}_1$ and $\mathcal{M}_2$. Operated through a coding agent with shell-level access, this agentic harness workflow executes a four-stage process to produce the benchmark $B^{(t)}$ and its validation- and audit-derived telemetry for subsequent optimization rounds. 
Overall, structured into two core phases, the pipeline balances generative creation with rigorous governance. The first part, encompassing \textit{Scenario Discovery} and \textit{Task \& Eval Synthesis}, systematically generates novel tasks paired with deterministic evaluators. Subsequently, the second part, comprising \textit{Model Validation} and \textit{Lifecycle Audit}, executes these tasks and governs their lifecycles to produce critical execution telemetry, feeding this data directly back into the outer loop of meta-harness orchestration to drive subsequent iterations. 
First, during \textit{Scenario Discovery}, the agent explores new problem directions for task evolution by mutating selected seed items into diverse candidate scenarios, with redundant copies and invalid replays excluded by the novelty filter. 
Second, building upon these candidate scenarios, \textit{Task \& Eval Synthesis} synthesizes concrete task families paired with shared deterministic evaluators, while an internal validity filter discards failing tasks. 
Third, \textit{Model Validation} executes the generated benchmark suite across the proxy models to capture responses, intermediate artifacts, and objective metrics, thereby producing the execution telemetry required by the outer loop optimization. 
Fourth, building directly upon these validation outcomes, the final stage, \textit{Lifecycle Audit}, assigns discrete scenario-level management states (\textsc{Keep-and-Evolve}, \textsc{Keep-Stable}, \textsc{Reseed}) to each scenario, sustainably steering the benchmark's evolution and maximizing search efficiency rather than simply discarding search budgets. 
Ultimately, this continuous cycle of generation, validation, and curation feeds directly into the meta-harness, driving progressive improvements across successive optimization rounds. 

\begin{algorithm}[t]
\caption{The \textsc{MetaBench-Harness} benchmark harness optimization procedure.}
\label{alg:search_loop}
\small
\begin{algorithmic}[1]
\REQUIRE Agent $\mathcal{M}_{\text{agent}}$, Engine $R_G$, Seed dataset $B_{\text{seed}}$, Proxy models ($\mathcal{M}_1, \mathcal{M}_2$), Max round $T$
\STATE \textbf{Initialize:} $H^{(0)}$; \ $B^{(0)} \leftarrow R_G(H^{(0)}, B_{\text{seed}})$; \ $\mathcal{T} \leftarrow \{(H^{(0)}, B^{(0)}, \text{ValidateAndAudit}(B^{(0)} \mid \mathcal{M}_1, \mathcal{M}_2))\}$
\FOR{$t=1 \dots T$}
    \STATE $\mathcal{M}_{\text{agent}}$ proposes candidates $\{\widetilde{H}_1,\dots,\widetilde{H}_{K}\}$ based on $H^{(t-1)}$ \hfill 
        $\triangleright$ Outer loop harness proposal 
    \FOR{$i = 1 \dots K$}
        \IF{$\widetilde{H}_i$ passes interface validation}
            \STATE $v_i \leftarrow \text{Evaluate}(\widetilde{H}_i \mid \mathcal{T})$ \hfill $\triangleright$ Harness proposal evaluation
        \ENDIF
    \ENDFOR
    \STATE $i^* \leftarrow \arg\max_i v_i; \ H^{(t)} \leftarrow \widetilde{H}_{i^*}$ \hfill $\triangleright$ Select best benchmark harness proposal
    \STATE $B^{(t)} \leftarrow \emptyset$
    \FOR{each scenario $s \in B^{(t-1)}$}
        \STATE $\quad s^{(t)} \leftarrow R_G(H^{(t)}, s, B_{\text{seed}})$ 
        \hfill $\triangleright$ Inner loop benchmark generation
        \STATE $\quad B^{(t)} \leftarrow B^{(t)} \cup \{s^{(t)}\}$
    \ENDFOR
    \STATE $r^{(t)} \leftarrow \text{ValidateAndAudit}(B^{(t)} \mid \mathcal{M}_1, \mathcal{M}_2)$; \ $\mathcal{T} \leftarrow \mathcal{T} \cup \{(H^{(t)}, B^{(t)}, r^{(t)})\}$ \hfill $\triangleright$ Validation and Audit
\ENDFOR
\RETURN $(B^{(T)}, H^{(T)})$
\end{algorithmic}
\end{algorithm}

\textbf{End-to-End Optimization Procedure and Deliverables. }
To clearly illustrate the dual-loop operation, we formalize the procedure as pseudocode in Algorithm~\ref{alg:search_loop}. The procedure begins by initializing the trajectory memory using a predefined benchmark harness and a seed dataset. The optimization then executes over $T$ rounds, with each round comprising two main stages: harness proposal and benchmark generation. During the harness proposal stage, the LLM agent generates $K$ candidate configurations, and an LLM-as-a-judge mechanism evaluates these candidates to select the optimal benchmark harness. In the subsequent generation stage, the inner loop leverages this selected harness and seed data to evolve prior scenarios, accumulating them into the updated benchmark suite. This suite then undergoes a \textit{ValidateAndAudit} phase, where proxy models execute the benchmark to provide empirical feedback for subsequent evolution. This validation score, alongside the round's harness and benchmark states, is appended to the trajectory memory to guide future search iterations. Upon completing all $T$ rounds, the procedure outputs the fully optimized evaluation benchmark and its corresponding harness configuration. Note that although the trajectory memory comprehensively logs all execution artifacts in practice, its representation is simplified here for clarity. 

\textbf{Module Instantiation of \sys. }
Within our framework, the core modules are operationalized by an LLM agent ($\mathcal{M}_{\text{agent}}$) guided by structured Markdown scaffolds. These templates explicitly define the expected functionality, inputs, outputs, and operational constraints for each distinct module. Furthermore, because our optimization objective is to discover a harness that yields challenging yet highly discriminative questions, we employ two distinct LLMs to serve as the underlying proxy models ($\mathcal{M}_1$ and $\mathcal{M}_2$) during the \texttt{ValidateAndAudit} phase. Specific model choices, system prompts, and detailed empirical setups are deferred to Section~\ref{sec:experiments} and Appendix~\ref{app:implementation_details_prompt}. 

\section{Experiments}
\label{sec:experiments}

In our implementation of \sys, we instantiate the central agent ($\mathcal{M}_{\text{agent}}$) with \texttt{Claude-Sonnet-5}~\citep{anthropic2026claude}, and employ both \texttt{Claude-Sonnet-5} and \texttt{GPT-5.4}~\citep{openai2026gpt54} as the proxy models ($\mathcal{M}_1$ and $\mathcal{M}_2$) to provide the necessary difficulty and discrimination signals during the search phase. Driven by this setup, we assess the framework's efficacy by evolving benchmarks in the domains of coding and mathematics. Specifically, we utilize CodeContests \citep{li2022competition} and AIME-2024 \citep{jia2024aime2024} as our foundational seed datasets, executing the optimization procedure for $T=10$ and $T=5$ rounds respectively, to yield the final suites denoted as MB-CodeContests and MB-AIME. Because the framework equips each newly generated task with a programmatic evaluator, a model's prediction is considered correct only upon passing this execution check; consequently, we adopt pass rate (accuracy) as our primary evaluation metric. 

To verify that the evolved benchmarks remain challenging and discriminative beyond the inner-loop proxy models, we evaluate several held-out models, including 
\texttt{DeepSeek-V3.1}~\citep{deepseekai2024deepseekv3technicalreport}, 
\texttt{Kimi-K2.5}~\citep{team2026kimi}, 
and \texttt{Gemini-2.5-Flash}~\citep{google_gemini_2_5_flash}, 
alongside the distilled reasoning models 
\texttt{DeepSeek-R1-Distill-Qwen-32B}~\citep{deepseekai2025deepseekr1incentivizingreasoningcapability} 
and 
\texttt{DeepSeek-R1-Distill-Qwen-7B}~\citep{deepseekai2025deepseekr1incentivizingreasoningcapability}. 
For fair comparison, all held-out evaluations use zero-temperature decoding under a uniform inference budget, enforced by a 32,000-token output cap and a 900-second timeout per call. Further details are in Appendix~\ref{app:exp_details}. 

\subsection{Main Result: Performance of Held-Out Models on Evolved Benchmarks}
\label{subsec:main_results}
We evaluate \sys\ in terms of evolution strategies (Table~\ref{tab:evolution_method}) and different evolution rounds (Table~\ref{tab:consolidated_results}), analyzing how different approaches push task complexity and discriminate model capabilities.
Additional model and experiment configurations are available in Appendix~\ref{app:heldout_model_evaluation}. 

\begin{table}[t]
\centering
\small
\setlength{\tabcolsep}{3pt}
\renewcommand{\arraystretch}{1.3}
\begin{threeparttable}
\caption{Model performance on benchmarks evolved via different evolution methods. }
\label{tab:evolution_method}
\vspace{-1em}
\begin{tabularx}{\linewidth}{l l *{5}{>{\centering\arraybackslash}X}}
    \toprule
    Evolution Methods & Note & \texttt{DeepSeek} & \texttt{Kimi} & \texttt{Gemini} & \texttt{Qwen-32B} & \texttt{Qwen-7B} \\
    \midrule
    \texttt{Baseline}  &  \hbox{AIME-2024}   & 0.900 & 0.900 & 0.833 & 0.700 & 0.533 \\
    \hdashline
    \texttt{TRACE}  &  Reported  & N/A   & N/A   & N/A   & 0.533  & 0.393  \\
    \texttt{AutoEvoEval} &  Reproduced  & 0.700 & 0.633 & 0.500 & 0.400  & 0.367  \\
    \hdashline
    \rowcolor{BaseGray}
    \texttt{Naïve Prompting} &  Ablation  & 1.000     & 1.000     & 1.000     & 0.875   &  0.750      \\
    \rowcolor{BaseGray}
    \texttt{Fixed Harness} &  Ablation  & 0.817 & 0.767 & 0.633 & 0.600 & 0.133  \\
    \hdashline
    \rowcolor{BaseGray}
    & & 0.322 & 0.544 & 0.144 & 0.089 & 0.044 \\[-0.5ex]
    \rowcolor{BaseGray}
    \multirow{-2}{*}{\sys} & \multirow{-2}{*}{Our} & \scriptsize $\pm 0.038$ & \scriptsize $\pm 0.217$ & \scriptsize $\pm 0.019$ & \scriptsize $\pm 0.019$ & \scriptsize $\pm 0.051$ \\
    \bottomrule
\end{tabularx}
\begin{tablenotes}
\scriptsize
\item[*] The specific versions of the held-out models evaluated here are \texttt{DeepSeek-V3.1}, \texttt{Kimi-K2.5}, \texttt{Gemini-2.5-Flash}, \texttt{DeepSeek-R1-Distill-Qwen-32B}, \texttt{DeepSeek-R1-Distill-Qwen-7B}.
\end{tablenotes}
\end{threeparttable}
\end{table}

\begin{table*}[t]
  \centering
  \caption{Model performance across evolved benchmarks. Models are grouped into reasoning (R), non-reasoning (NR), and adaptive (AR) types. ``Involve.'' denotes whether the LLM was involved in the evolution loop; ``Base.'' is the seed benchmark, with ``Early/Late'' as different evolution rounds. }
  \vspace{-0.5em}
  \label{tab:consolidated_results}
  \small
  \setlength{\tabcolsep}{8.5pt}
  \renewcommand{\arraystretch}{1.3}
  \begin{tabular}{l cc ccc ccc}
  \toprule
 \multirow{2}{*}{Model} & \multirow{2}{*}{Type}  & \multirow{2}{*}{Loop}  & \multicolumn{3}{c}{MB-CodeContests} & \multicolumn{3}{c}{MB-AIME} \\
  \cmidrule(r{3pt}){4-6} \cmidrule(l{3pt}){7-9}
  &  &  & Base. & Early & Late & Base. & Early & Late \\
  \midrule
  \rowcolor{TopGray}
  \texttt{Claude-Sonnet-5} & AR  & \checkmark & 0.835 & 0.948 & 0.839 & 0.733 & 0.750 & 0.633 \\
  \rowcolor{TopGray}
  \texttt{GPT-5.4}         & NR & \checkmark & 0.586 & 0.591 & 0.293 & 0.067 & 0.125 & 0.100 \\
  \hdashline
  \texttt{DeepSeek-V3.1}    & NR & $\times$ & 0.736 & 0.845 & 0.396 & 0.667 & 0.125 & 0.133 \\
  \texttt{Kimi-K2.5}        & NR & $\times$ & 0.694 & 0.834 & 0.739 & 0.833 & 0.333 & 0.167 \\
  \texttt{Gemini-2.5-Flash} & NR & $\times$ & 0.458 & 0.484 & 0.278 & 0.767 & 0.167 & 0.100 \\
  \hdashline
  \texttt{DeepSeek-V3.1}    & R & $\times$ & 0.788 & 0.946 & 0.770 & 0.900 & 0.625 & 0.333 \\
  \texttt{Kimi-K2.5}        & R & $\times$ & 0.935 & 0.948 & 0.852 & 0.900 & 0.667 & 0.533 \\
  \texttt{Gemini-2.5-Flash}  & R & $\times$ & 0.799 & 0.728 & 0.585 & 0.833 & 0.167 & 0.167 \\
  \bottomrule
  \end{tabular}
\end{table*}

\textbf{Model Performance Across Different Evolution Methods.} 
We compare model performance across benchmark variants evolved from the AIME-2024 dataset, primarily focusing on \sys\ and its two ablated baselines, alongside AutoEvoEval \citep{wu2025autoevoeval} and TRACE \citep{guo2026towards_trace}. 
The \texttt{Baseline} refers to the original, unevolved AIME-2024 evaluation dataset. 
\texttt{Naïve Prompting} consists solely of \textit{Scenario Discovery} and \textit{Task \& Eval Synthesis} (Figure~\ref{fig:pipeline}b), omitting execution feedback entirely. 
\texttt{Fixed Harness} omits the outer-loop meta-harness orchestration (Figure~\ref{fig:pipeline}a) and relies only on a static harness (Figure~\ref{fig:pipeline}b). 
For \texttt{AutoEvoEval}, we select code- and math-related execution units for benchmark evolution, whereas for TRACE, we report their published results. 
Evaluation results summarized in Table~\ref{tab:evolution_method} highlight the distinct impacts of these design choices. Specifically, while \texttt{Baseline} evaluations suffer from severe saturation (yielding up to $0.900$ accuracy on \texttt{DeepSeek}), alternative methods offer only marginal difficulty scaling. 
\texttt{Naïve Prompting} even exacerbates this issue: without execution feedback to verify viability, generation collapses into trivial or flawed tasks where top models score a perfect $1.000$. In contrast, \sys\ systematically drives down performance across all models, dropping \texttt{DeepSeek} to $0.322$ and \texttt{Qwen-7B} to $0.044$. This demonstrates that both inner-loop execution feedback and outer-loop meta-harness orchestration are essential to eliminate performance saturation and expose fine-grained model capabilities.

\textbf{Model Performance Across Different Evolution Rounds.} 
We evaluate proxy (in-loop) and held-out models across \sys's base, early, and late evolution rounds. As shown in Table~\ref{tab:consolidated_results}, while proxy models typically peak early and decline as late-stage complexity increases, this evolution ultimately enhances benchmark quality: it better distinguishes model capabilities on MB-CodeContests, and escalates difficulty on MB-AIME while preserving strong discriminative power. 
For held-out models, late-stage difficulty scaling highlights a stark contrast between non-reasoning (NR) and reasoning (R) models. On the highly challenging MB-AIME, NR models suffer a near-total collapse, dropping to a narrow $0.100$--$0.167$ accuracy band. Conversely, R models demonstrate strong resilience; top reasoners like \texttt{Kimi-K2.5} and \texttt{DeepSeek-V3.1} maintain scores of $0.533$ and $0.333$, respectively. 
Finally, Table~\ref{tab:token_length_both_datasets} substantiates this increased complexity through reasoning lengths. By the late stage on MB-AIME, \texttt{Kimi-K2.5} and \texttt{DeepSeek-V3.1} require significantly more computational effort, generating an additional $14,764$ and $5,536$ tokens, respectively. 
Ultimately, the preserved discriminative power and increased reasoning effort prove that \sys\ generates genuinely reasonable benchmarks that test advanced models, rather than merely encouraging unintended shortcuts or flawed heuristics.

\begin{table*}[t]
\centering
\small
\caption{Comparison of model reasoning lengths under the base and late evolution datasets.}
\vspace{-0.5em}
\label{tab:token_length_both_datasets}
\setlength{\tabcolsep}{7.5pt}
\begin{tabular}{l c cc cc}
\toprule
\multirow{2}{*}{Model} & \multirow{2}{*}{Type} & \multicolumn{2}{c}{MB-CodeContests} & \multicolumn{2}{c}{MB-AIME} \\
\cmidrule(r{3pt}){3-4} \cmidrule(l{3pt}){5-6}
& & $\text{Base} \rightarrow \text{Late}$ & $\Delta$ Tokens & $\text{Base} \rightarrow \text{Late}$ & $\Delta$ Tokens \\
\midrule
\texttt{DeepSeek-V3.1}    & R & $11{,}223 \rightarrow 13{,}382$ & $(+2{,}159)$   & $10{,}208 \rightarrow 15{,}744$ & $(+5{,}536)$ \\
\texttt{Kimi-K2.5}        & R & $15{,}879 \rightarrow 16{,}732$ & $(+853)$   & $14{,}218 \rightarrow 28{,}982$ & $(+14{,}764)$ \\
\texttt{Gemini-2.5-Flash} & R & $15{,}537 \rightarrow 16{,}562$ & $(+1{,}025)$ & $11{,}254 \rightarrow 17{,}020$ & $(+5{,}766)$ \\
\bottomrule
\end{tabular}
\end{table*}

\begin{table*}[t]
\centering
\begin{minipage}[t]{0.54\textwidth}
\centering
\small
\setlength{\tabcolsep}{3pt}
\renewcommand{\arraystretch}{1.2}
\caption{Harness-Level Evolutionary Changes.}
\vspace{-0.5em}
\label{tab:meta_change_stats}
\begin{tabular*}{\linewidth}{@{\extracolsep{\fill}} lcc @{}}
\toprule
& MB-CodeContests & MB-AIME \\
\midrule
\rowcolor[HTML]{EFEFEF}
\multicolumn{3}{l}{\emph{Meta-Harness Orchestrated Proposals}} \\
\quad Optimization rounds & 10 & 5 \\
\quad Proposals applied & 20/23 (87\%) & 10/12 (83\%) \\
\midrule
\rowcolor[HTML]{EFEFEF}
\multicolumn{3}{l}{\emph{Edits Applied by the Harness}} \\
\quad Measurement  & 11 (39\%) & 2 (20\%) \\
\quad Generation   & 9 (32\%) & 6 (60\%) \\
\quad Execution   & 5 (18\%) & 2 (20\%) \\
\quad Telemetry   & 3 (11\%) & 0 (0\%) \\
\bottomrule
\end{tabular*}
\end{minipage}
\hfill
\begin{minipage}[t]{0.42\textwidth}
\centering
\small
\setlength{\tabcolsep}{2pt}
\renewcommand{\arraystretch}{1.2}
\caption{Benchmark Evolution Statistics.}
\vspace{-0.5em}
\label{tab:benchmark_stats}
\begin{tabular*}{\linewidth}{@{\extracolsep{\fill}} lcc @{}}
    \toprule
         & \#. Ori. & \#. Evo. \\
    \midrule
    \rowcolor[HTML]{EFEFEF}
    \multicolumn{3}{l}{\textit{MB-CodeContests}} \\
    \quad Total Questions & 165 & 100 \\
    \quad Solving Techniques   & 5.02 & 5.49 \\
    \quad Capability & 3.55 & 3.90 \\
    \midrule
    \rowcolor[HTML]{EFEFEF}
    \multicolumn{3}{l}{\textit{MB-AIME}} \\
    \quad Total Questions & 30 & 30 \\
    \quad Solving Techniques & 3.67 & 4.73 \\
    \quad Capability & 3.30 & 3.40 \\
    \bottomrule
\end{tabular*}
\end{minipage}
\end{table*}

\subsection{Benchmark Harness Evolution Analysis and Data Quality Assessment}
\label{subsec:harness_evolution}
 
The previous section demonstrated that \sys\ effectively escalates benchmark difficulty while preserving strong model discriminability. Building on these findings, this section examines the underlying dynamics of the evolution procedure, assesses the quality of the generated benchmark, and analyzes the specific model capabilities required by the evolved tasks.

\textbf{How does the meta-harness adapt to different domains? }
As shown in Table~\ref{tab:meta_change_stats}, our framework maintains a high acceptance rate for harness proposals across both coding and math scenarios. However, the meta-harness reveals distinct adaptation trends tailored to each domain's inherent challenges.
For coding tasks in CodeContests, the system heavily modifies \textit{Measurement} (39\%). Unlike math, programming tasks cannot be judged by a single final answer alone; they require different unit tests, edge-case checks, and performance tracking. Consequently, the meta-harness frequently updates measurement tools to ensure rigorous evaluation. Furthermore, code execution generates rich intermediate data, necessitating robust \textit{Telemetry} edits (11\%) to capture execution logs, runtime errors, and trace signals for subsequent evolution adjustments.
Conversely, math task evolution in MB-AIME concentrates primarily on \textit{Generation} (60\%). Because AIME problems have clear, deterministic numerical answers, the evaluation mechanism is straightforward, placing minimal burden on measurement. Instead, the core challenge lies in evolving the task itself. 
The meta-harness directs most of its modifications toward task generation and structural mutation, continuously altering constraints and problem settings to prevent models from relying on memorized templates and to push the limits of advanced mathematical reasoning. 
Lastly, the details of Table~\ref{tab:meta_change_stats} can be found in Appendix~\ref{app:harness_edits}, and a round-by-round qualitative analysis is provided in Appendix~\ref{app:qualitative_analysis_and_case_study}. 

\begin{figure*}[t]
    \centering
    \includegraphics[width=\textwidth]{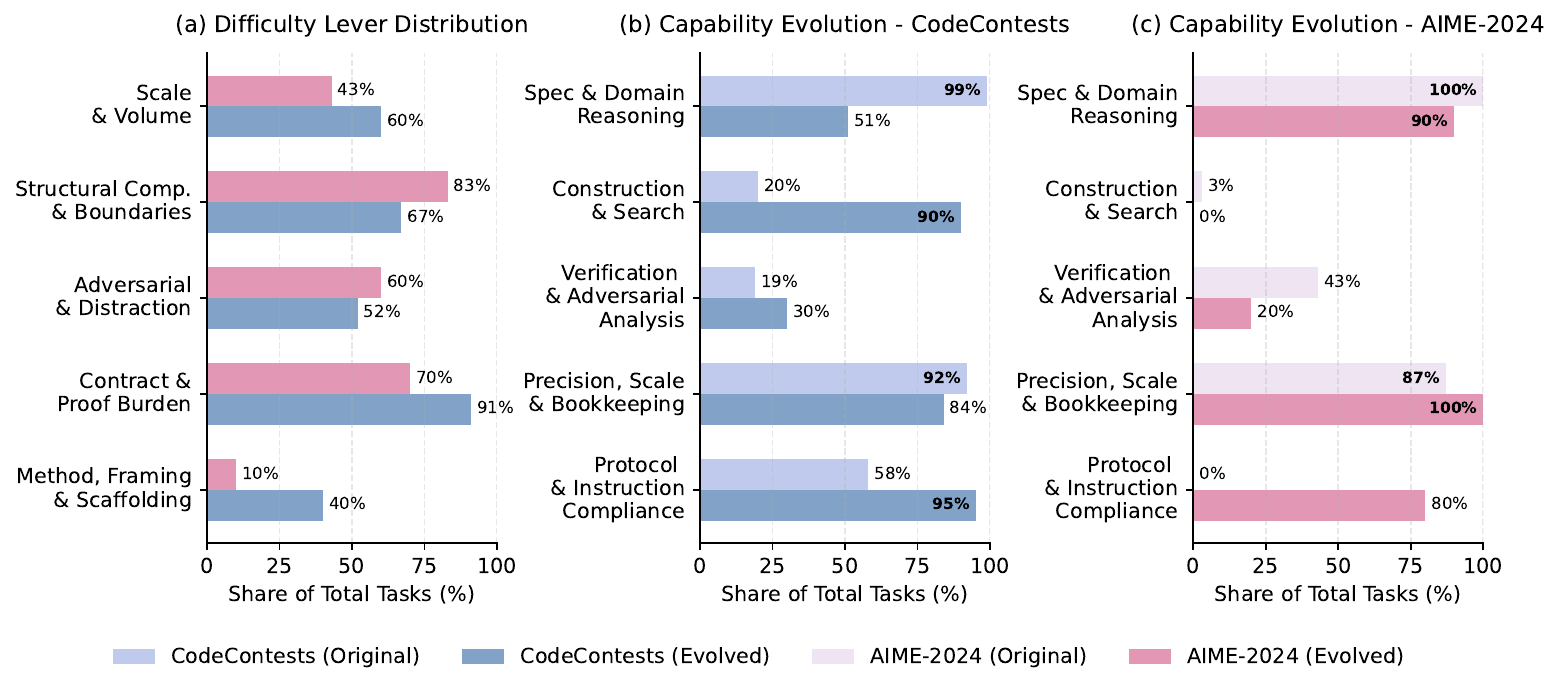}
    \vspace{-2em}
    \caption{Quantitative analysis of evolved benchmarks across MB-CodeContests and MB-AIME. 
    }
    \label{fig:macro_taxonomy_all}
\end{figure*}

\textbf{Which dimensions of the benchmark's capabilities are evolved? }
Having detailed the mechanical edits applied by the meta-harness, this section now examines how these changes fundamentally transform the tasks within the evolved benchmarks. 
Table~\ref{tab:benchmark_stats} compares the original datasets with their counterparts evolved by \sys. 
Across both domains, the evolved tasks demand more capability stages, indicating a substantial increase in overall task complexity. 
Evolution elevates task complexity by engaging more capability stages across the solving process. 
To analyze how task complexity scales, Figure~\ref{fig:macro_taxonomy_all} utilizes \texttt{Claude-Fable-5} \citep{anthropic2026fablemythos} to compare evolved tasks with their original seed questions across two dimensions (difficulty levers and capability dimensions). By sampling a subset of examples for human validation, we confirm high Human--AI agreement of 97.1\% and 96.1\% for the two dimensions, respectively.
Figure~\ref{fig:macro_taxonomy_all} (a) shows that both domains evolve primarily by tightening answer criteria and compounding constraints. While mathematical tasks lean toward structural composition, coding tasks increase the scale of required work. Figures~\ref{fig:macro_taxonomy_all} (b) and (c) further illustrate the shift in required capabilities following evolution across the two datasets. Mathematical tasks emphasize precise computation, multi-case bookkeeping, and structured output compliance, while coding tasks shift toward constructive solution generation under strict evaluation constraints. 
See Appendix~\ref{app:breakdown_analysis} for a detailed breakdown. 

\textbf{Does the benchmark quality improve through evolution? }
To ensure the evolved benchmarks are meaningful, we pairwise compare evolved tasks against their seeds using a more powerful LLM judge, \texttt{Claude-Opus-5} \citep{anthropic2026claudeopus}, as summarized in Table~\ref{tab:validity_audit}. 
\begin{wraptable}[15]{r}{0.48\textwidth} 
\centering
\small
\setlength{\tabcolsep}{3pt} 
\renewcommand{\arraystretch}{1.2}
\caption{Evolution Quality Assessment Results: Final and overall average success rates. ``Human'' denotes the Human–AI Agreement, measuring consistency between human and agent.}
\vspace{-0.5em}
\label{tab:validity_audit}
\begin{tabular}{lccc}
\toprule
Metric & MB-CC$^{*}$ & MB-AIME & \makecell{Human} \\
\midrule
Reasonableness & 95\%\,/\,91\% & 90\%\,/\,83\% & 85\% \\
Competency     & 78\%\,/\,61\% & 80\%\,/\,78\% & 68\% \\
Robustness     & 90\%\,/\,62\% & 93\%\,/\,69\% & 85\% \\
\hdashline
Overall        & 70\%\,/\,44\% & 70\%\,/\,49\% & 80\% \\
\bottomrule
\multicolumn{4}{l}{\scriptsize \emph{$^*$MB-CC denotes MB-CodeContests.}} \\
\end{tabular}
\end{wraptable}
For each dimension, the judge outputs a binary validation, with the reported success rates reflecting the proportion of tasks that meet the criteria. 
To validate the reliability of the LLM judge, we also measure Human--AI agreement on a sampled subset (reported as the Human column). 
First, \textit{Reasonableness} evaluates whether the evolved task remains valid and well-formed, preserving a recognizable lineage while introducing meaningful changes. 
Second, \textit{Competency} verifies that the task faithfully measures the intended capability, ensuring that it demands genuine reasoning rather than adding superficial complexity. 
Lastly, \textit{Robustness} examines whether the evolved evaluator can reliably distinguish valid from invalid solutions, maintaining strict resistance to formatting artifacts, implementation errors, and adversarial exploitation. 
Crucially, the \textit{Overall} metric demands simultaneous success across all three dimensions, with this joint pass rate in the final round substantially outperforming the cross-round average. 
This consistent upward trend demonstrates that our iterative pipeline progressively converges toward highly valid and structurally robust tasks. (See Appendix~\ref{app:quality_assessment} for details. )

\begin{table*}[t]
\centering
\small
\caption{\sys\ Evolution Case Study: From Initial Seed to Deepening}
\vspace{-0.8em}
\label{tab:merged_evolution_case_study}

\begin{tabularx}{\textwidth}{>{\RaggedRight\hsize=1.0\hsize}X >{\RaggedRight\hsize=1.0\hsize}X >{\RaggedRight\hsize=1.0\hsize}X}
\multicolumn{3}{c}{Part I: Case Study of MB-CodeContests Evolution} \\
\toprule
\textbf{Seed} (Codeforces 1582D) & \textbf{Round 5} (Evolved Emergence) & \textbf{Round 9} (Evolved Deepening) \\
\midrule

\rowcolor{gray!20}
\multicolumn{3}{l}{\textit{Task Formulation}} \\
\addlinespace[2pt]
\textbf{Constructive Math}: Given an array $a$ ($a_i \neq 0$), construct $b$ such that $\sum a_i b_i = 0$ and $\sum |b_i| \le 10^9$. & 
\textbf{Single Gap Audit}: Exploit a planted gap (\texttt{range(n-1)}) to trick a Python checker with an invalid $(a, b)$ pair. &
\textbf{Compound Audit}: Trace complex logic to exploit a vulnerability gated by 3 conditions ($n > 1000$, duplicate $a$, odd $n$). \\
\addlinespace[4pt]

\midrule
\rowcolor{gray!20}
\multicolumn{3}{l}{\textit{Evolution Summary}} \\
\addlinespace[2pt]
\textbf{Capability:} 3 \quad
\textbf{Technique:} 3 \par\smallskip
\textbf{Lever:} Baseline specification \par\smallskip
\textbf{Baseline:} The root foundation originating from a standard coding task.
&
\textbf{Capability:} 3 \quad
\textbf{Technique:} 5 \par\smallskip
\textbf{Key Lever:} Meta level shift, Output contract tightening \par\smallskip
\textbf{Moderate:} Shifts to basic code auditing and static inspection.
&
\textbf{Capability:} 4 \quad
\textbf{Technique:} 6 \par\smallskip
\textbf{Key Lever:} Compound conjunction, Scaffolding removal \par\smallskip
\textbf{High:} Mandates multi-technique fusion without hints. \\
\bottomrule\end{tabularx}

\vspace{0.5em} 

\begin{tabularx}{\textwidth}{>{\RaggedRight\hsize=1.0\hsize}X >{\RaggedRight\hsize=1.0\hsize}X >{\RaggedRight\hsize=1.0\hsize}X}
\multicolumn{3}{c}{Part II: Case Study of MB-AIME Evolution} \\
\toprule
\textbf{Seed} (AIME 2024-II-9) & \textbf{Round 1} (Evolved Emergence) & \textbf{Round 5} (Evolved Deepening) \\
\midrule

\rowcolor{gray!20}
\multicolumn{3}{l}{\textit{Task Formulation}} \\
\addlinespace[2pt]
\textbf{Combinatorial Counting}: Count placements on a $5 \times 5$ grid where no further chip can be added (\emph{answer}: $902$). &
\textbf{Pre-Blocked Obstacle}: Count maximal placements on a $4 \times 4$ board with 2 permanently blocked cells (\emph{answer}: $150$). &
\textbf{Post-Hoc Filter}: Count maximal placements on a $3 \times 5$ board, then filter out those hitting 4 flagged cells (\emph{answer}: $14$). \\
\addlinespace[4pt]

\midrule
\rowcolor{gray!20}
\multicolumn{3}{l}{\textit{Evolution Summary}} \\
\addlinespace[2pt]
\textbf{Capability:} 2 \quad 
\textbf{Technique:} 4 \par\smallskip
\textbf{Lever:} Baseline specification \par\smallskip
\textbf{Baseline:} The root foundation originating from a standard olympiad task.
&
\textbf{Capability:} 3 \quad
\textbf{Technique:} 3 \par\smallskip
\mbox{\textbf{Key Lever:} Compound conjunction} \par\smallskip
\textbf{Moderate:} Forms compound constraints by adding permanent obstacles to maximal placement. 
&
\textbf{Capability:} 3 \quad
\textbf{Technique:} 4 \par\smallskip
\textbf{Key Lever:} Distractor injection\par\smallskip
\textbf{High:} Requires a multi-step deduction to filter out injected distractors without hints. \\
\bottomrule
\end{tabularx}
\end{table*}

\textbf{From Seed to Spark: Case Studies.} 
Table~\ref{tab:merged_evolution_case_study} shows two task evolution trajectories that highlight our core evolution pattern: the benchmark harness evolved by \sys\ systematically incorporates difficulty levers to drive multi-dimensional evolution across required capabilities and solving techniques. 
In Part I (MB-CodeContests), the system transforms the problem type, shifting from a standard constructive math seed to a complex code-auditing task. 
By applying levers like \textit{meta level shift} and \textit{scaffolding removal}, it forces multi-technique fusion, expanding capabilities to include static code analysis. 
Similarly, Part II (MB-AIME) evolves a counting problem via \textit{compound conjunction} and \textit{distractor injection}, shifting the required skills from pure domain reasoning to include instruction compliance and distractor resistance. 
Rather than merely scaling difficulty along a single axis, the evolved harness uses diverse levers to formulate questions that demand additional capabilities and solving techniques. 
Detailed case studies are in Appendix~\ref{app:appendix_case_study}. 

\section{Conclusion}
\label{sec:conclusion}
In this paper, we propose \textsc{MetaBench-Harness}, a dual-loop framework designed for the end-to-end optimization of benchmark generation workflows. 
By combining an inner loop for task generation with an outer meta-harness layer that refines harness implementations based on historical trajectories, we transform benchmark creation from a static process into a dynamic evolution. 
Using CodeContests and AIME-2024 as seed tasks, our experiments show that the framework generates benchmarks that are both challenging and discriminative for models. 
Supported by trajectory and quality analyses, we verify that it enables multi-dimensional evolution, and steadily improves evolution reasonableness, benchmark competency, and evaluator robustness across successive evolution rounds. 
Furthermore, case studies quantitatively show that it not only increases the required capabilities and problem-solving techniques, but also uses different difficulty levers to reframe the problems. 
More broadly, this work provides a solution to static benchmark saturation. 
\section*{Reproducibility Statement}

To ensure high reproducibility, our pipeline automatically generates a matching code-based evaluator for every new task. This guarantees that future evaluations on our dataset remain consistent and fair, avoiding the instability of subjective LLM-as-a-judge scoring. Furthermore, all seed datasets, prompt templates, and the complete codebase will be publicly released upon paper acceptance. 

\section*{The Use of Large Language Models}
In this work, Large Language <odels (LLMs) play multiple roles.
First, as the core module of our research, they act as agents within the \textsc{MetaBench-Harness} framework for autonomous task generation, evaluator synthesis, and meta-harness orchestration, with their generation quality rigorously assessed through human validation.
Second, they serve as the target models evaluated in our experiments.
Third, during manuscript preparation, we utilized LLMs strictly as writing assistants. This usage was strictly limited to language refinement, including grammar correction, clarity improvement, and avoiding ambiguous expressions, without generating any scientific claims or core ideas.
We explicitly emphasize that all scientific claims, experimental designs, logical arguments, and core contributions were entirely conceived and proposed by human authors in this paper.

\bibliography{iclr2027_conference}
\bibliographystyle{iclr2027_conference}

\newpage
\appendix
\section*{Appendix}

\section{The Implementation Details of \sys}
\label{app:implementation_details_prompt}
This section details the concrete implementation and module design of \sys. Expanding upon the module instantiation introduced in Section~\ref{sec:design_details}, we explain how the primary LLM agent ($\mathcal{M}_{\text{agent}}$) operationalizes the structured Markdown scaffolds. We begin with a high-level overview of the system's core mechanics, covering Master Initialization, Module Instantiation, and the overarching Dual-Loop Evolution workflow. 

\noindent\textbf{Master Initialization.} Serving as the foundational entry point, this initial setup establishes the operational constraints and global configuration for a \sys\ run. Executed by $\mathcal{M}_{\text{agent}}$ (operating as a coding agent with shell access) through a single master configuration file, it specifies the seed dataset, item formatting conventions, overarching objectives, stage sequencing, dual-loop workflow, and the mandatory on-disk locations for all generated artifacts and scripts. 

\noindent\textbf{Module Instantiation.} Guided by the master configuration, $\mathcal{M}_{\text{agent}}$ executes the workflow's distinct modules by sequentially traversing the Markdown scaffolds. Treating each prompt as a unit of work, the agent reads prior artifacts, executes its objective, and writes the required outputs (e.g., JSON, Python) to mandated locations. A stage is complete only when these files exist on disk; this rigorous file-passing discipline, rather than a persistent context, robustly carries state through the round. 

\noindent\textbf{Dual-Loop Evolution.} 
This self-evolving workflow organizes the system's modules into two alternating cycles. 
The outer loop drives the system's meta-optimization, where $\mathcal{M}_{\text{agent}}$ analyzes archived telemetry (e.g., score distributions and failure analyses) to propose, evaluate, and apply edits directly to the benchmark harness's Markdown scaffolds, effectively rewriting its underlying instructions. 
Under this updated setup, the inner loop acts as the execution engine for the harness itself. It runs the revised sequence of generation and measurement modules to generate the evolved benchmarks, while collecting fresh telemetry for the next update. 
This benchmark evolutionary cycle repeats until the stop controller terminates the execution. 

To unpack the \sys\ workflow, the following subsections detail its three primary phases: Master Initialization (initial setup), Meta-Harness Orchestration (outer loop), and Benchmark Generation (inner loop). To highlight the core logic of each module, the Markdown scaffolds presented below are condensed summaries rather than exhaustive full prompts. 

\subsection{Master Initialization}
\label{sec:app_master}
This phase establishes the global constraints and foundational anchors for the entire execution. The agent first ingests the master configuration to strictly bind the pipeline to the target dataset and define the baseline evaluation thresholds for all subsequent inner-loop operations. 

\begin{stepbox} 
--- Master Initialization --- \hfill \texttt{[00\_master\_start.md]}
\begin{enumerate}[leftmargin=1.5em, itemsep=2pt, topsep=4pt]
    \item \textbf{Seed Anchor}: \texttt{pipeline.yaml} and \texttt{00\_master\_start.md} solely define dataset-level facts; no other prompt may hard-code seed details. 
    \item \textbf{Evolution Rule}: every scenario and task must be seed-derived with explicit \texttt{seed\_lineage}; transformation, never copying.
    \item \textbf{Difficulty Defaults}: too-easy task (mean of at least 0.85 or universal pass); too-easy scenario (mean of at least 0.80, failure rate of at most 0.20); non-discriminative (spread of at most 0.10); healthy band between 0.25 and 0.65. 
    \item \textbf{Hard Constraints}: deterministic programmatic verification per task; strict artifact/script layout (\texttt{data/rounds/\{id\}/}, \texttt{scripts/}).
\end{enumerate}
\end{stepbox}

\subsection{Outer Loop: Meta-Harness Orchestration}
\label{sec:app_meta_harness}

Following Section~\ref{sec:design_details}, we detail the meta-harness that \emph{evolves} the harness. Its four prompts operate exclusively on archived round telemetry instead of raw model answers, keeping optimization strictly decoupled from individual evaluations to ensure robust and unbiased self-improvement.

\subsubsection{Stages 1: Telemetry Review and Harness Proposer}

\begin{stepbox}
--- Proposal Workflow (shared) --- \hfill \texttt{[mh/01\_harness\_proposer.md]}
\begin{enumerate}[leftmargin=1.5em, itemsep=2pt, topsep=4pt]
    \item \textbf{Telemetry Review}: read the round's summaries, score distributions, failure analyses, rejection logs, and lifecycle decisions. 
    \item \textbf{Candidate Generation}: emit $3$--$5$ scoped candidate revisions to specific stage prompts or configs (difficulty escalation, evaluator reliability, lifecycle policy), each with evidence, expected score direction, risks, and a rollback plan. 
    \item \textbf{Seed Guardrails}: preserve seed lineage, deterministic scoring, and \texttt{00\_master\_start.md} as the single source of truth. 
    \item \textbf{Output}: \texttt{proposal.md}/\texttt{.json} per candidate, plus an index. 
\end{enumerate}
\end{stepbox}

\subsubsection{Stage 2: Harness Evaluator}
\begin{stepbox}
--- Evaluation Workflow (shared) --- \hfill \texttt{[mh/02\_harness\_evaluator.md]}
\begin{enumerate}[leftmargin=1.5em, itemsep=2pt, topsep=4pt]
    \item \textbf{Ten-Dimension Scoring}: rate each candidate 1--5 on difficulty pressure, discrimination, evaluator reliability, task validity, novelty, seed-evolution fidelity, lifecycle quality, reproducibility, cost control, and implementation risk.
    \item \textbf{Hard Rejection}: reject candidates that weaken deterministic verification, remove seed-lineage requirements, hard-code seed items, enable from-scratch generation, or add difficulty via ambiguity or brittle formatting.
    \item \textbf{Weighted Judgment}: compute \texttt{overall\_harness\_score} via weights (difficulty pressure, evaluator reliability, task validity, seed fidelity, reproducibility weighted highest), not a blind average. 
    \item \textbf{Output}: per-candidate \texttt{evaluation.json}/\texttt{.md} and a summary. 
\end{enumerate}
\end{stepbox}

\subsubsection{Stage 3: Harness Selector} 
\begin{stepbox}
--- Selection Workflow (shared) --- \hfill \texttt{[mh/03\_harness\_selector.md]}
\begin{enumerate}[leftmargin=1.5em, itemsep=2pt, topsep=4pt]
    \item \textbf{Ranking}: aggregate evaluation results with the persistent cross-round \texttt{leaderboard.json}. 
    \item \textbf{Selection Rule}: select at most one candidate per round, unless two are complementary and low-risk; selecting none is valid. 
    \item \textbf{Invariant Check}: re-verify that the selected candidate preserves the seed-evolution contract and the configuration source of truth.
    \item \textbf{Output}: updated leaderboard, \texttt{selected\_candidate.json}, and a report.
\end{enumerate}
\end{stepbox}

\subsubsection{Stage 4: Harness Applier}
\begin{stepbox}
--- Application Workflow (shared) --- \hfill \texttt{[mh/04\_harness\_applier.md]}
\begin{enumerate}[leftmargin=1.5em, itemsep=2pt, topsep=4pt]
    \item \textbf{Scoped Edits}: apply the selected revision directly to live files (e.g., stage prompts~$\mathcal{P}$, \texttt{pipeline.yaml} / difficulty configuration~$\mathcal{L}$, or lifecycle policy~$\Pi_{\text{life}}$), strictly limited to those in \texttt{selected\_candidate.json}. 
    \item \textbf{Integrity Check}: confirm the seed-evolution contract and master configuration remain intact after editing.
    \item \textbf{No-Op Reporting}: if nothing was selected, write an apply report stating that the next round runs under the unchanged harness.
    \item \textbf{Output}: \texttt{apply\_report.md}/\texttt{.json} enumerating every modified file, so that round $t{+}1$ verifiably runs on $\mathcal{H}^{(t+1)}$.
\end{enumerate}
\end{stepbox}

\subsection{Inner Loop: Benchmark Generation}
\label{sec:app_benchmark_gen}
Having specified how the harness \emph{evolves}, we now detail the inner-loop pipeline that each harness state instantiates to generate a round's benchmark. This execution strictly follows a four-stage sequence: Scenario Discovery, Task \& Evaluator Synthesis, Model Validation, and a Lifecycle Audit. 

\subsubsection{Stage 1: Scenario Discovery}
\begin{stepbox}
--- Discovery Workflow (shared) --- \hfill \texttt{[01\_scenario\_discovery.md]}
\begin{enumerate}[leftmargin=1.5em, itemsep=2pt, topsep=4pt]
    \item \textbf{Seed Selection}: load the configured dataset or local manifest; select one seed item supporting deterministic evolution; save \texttt{selected\_seed\_item.json} and a seed-selection report.
    \item \textbf{Over-Generation}: produce $(B-K)\times\mu$ candidate scenarios ($100$ in round~1, where $K{=}0$), each a capability-centered seed evolution. 
    \item \textbf{Lineage Contract}: every candidate carries \texttt{seed\_lineage} and an explicit account of why it is a transformation, not a copy.
    \item \textbf{Output}: candidate JSONs under \texttt{scenarios\_raw/}.
\end{enumerate}
\end{stepbox}

\begin{stepbox}
--- Filtering and Selection Workflow (shared) --- \hfill \texttt{[02-03\_*.md]}
\begin{enumerate}[leftmargin=1.5em, itemsep=2pt, topsep=4pt]
    \item \textbf{Lineage Validation}: confirm seed lineage; flag \texttt{seed\_copy\_risk}; score novelty, feasibility, evaluator reliability, and degree of seed evolution (1--5 each); write \texttt{novelty\_reports/\{scenario\_id\}.json} with a \texttt{keep}/\texttt{revise}/\texttt{drop} recommendation.
    \item \textbf{Slot Refill}: carry over kept scenarios; select only enough survivors to restore the active budget $B{=}10$.
    \item \textbf{Diversity Enforcement}: the selected set must cover distinct transformation types, not near-duplicate variants of one.
    \item \textbf{Output}: \texttt{scenarios\_selected/} plus a selection report documenting diversity and seed-evolution rationale.
\end{enumerate}
\end{stepbox}

\begin{domainbox}
\textbf{Domain instantiation.} The CodeContests novelty checker
additionally enumerates valid programming-domain transformation axes
(reasoning burden, edge-case distribution, testing strategy, output
contract, solution-debugging objective) as acceptable evolutions of a
competitive-programming seed.
\end{domainbox}

\subsubsection{Stage 2: Task and Evaluator Synthesis}

\begin{stepbox}
--- Task and Evaluator Synthesis Workflow (shared) --- \hfill \texttt{[04-06\_*.md]}
\begin{enumerate}[leftmargin=1.5em, itemsep=2pt, topsep=4pt]
    \item \textbf{Difficulty-Aware Generation}: for families whose previous mean~$\ge 0.85$ or all models passed, same-difficulty siblings forbidden; escalation only via validity-preserving levers (constraint composition, counterfactuals, boundary cases), never ambiguity. 
    \item \textbf{Evaluator Construction}: emit \texttt{eval.py} implementing \texttt{evaluate(model\_output, item)}; execute three validation probes (ground-truth must pass, wrong-answer must fail, malformed-output must raise a format error) and assign a reliability grade; grades below $4$ are unusable. 
    \item \textbf{QA Gates}: accept only if the evaluator (validated \texttt{eval.py}, reliability~$\ge 4$), seed-evolution (valid lineage, low copy risk), and empirical-difficulty (reject near-equivalents of previous tasks with mean $\ge 0.85$ or universal pass) gates all hold. 
    \item \textbf{Output}: survivors in \texttt{tasks\_filtered/}, rejections logged with reasons, evaluators in \texttt{scripts/evaluators/}.
\end{enumerate}
\end{stepbox}
\vspace{-0.5em}

\begin{domainbox}
\textbf{Domain instantiation (AIME-2024).} Each math task requires an \texttt{answer\_verification} object: an agent-written deterministic checker must reproduce the expected answer with run evidence. Ground truth that cannot be re-derived is rejected at QA as evaluator-fragile. CodeContests needs no such object, as its evaluator directly checks correctness.
\end{domainbox}

\subsubsection{Stage 3: Model Validation}

\begin{stepbox}
--- Execution Workflow (shared) --- \hfill \texttt{[07\_model\_runner.md]}
\begin{enumerate}[leftmargin=1.5em, itemsep=2pt, topsep=4pt]
    \item \textbf{Reproducible Script}: generate a per-round runner under \texttt{scripts/model\_runners/}; load models and credentials from \texttt{models.yaml} and \texttt{.env} via safe loaders; never print or persist secrets.
    \item \textbf{Token-Budget Discipline}: request the largest safe per-model output budget; retry once at a lower accepted value on rejection.
    \item \textbf{Truncation Handling}: detect token-limit finish reasons, continue once when safely supported, and record truncation explicitly rather than silently scoring incomplete answers.
    \item \textbf{Output}: structured JSONL outputs with token-limit metadata, plus an execution report.
\end{enumerate}
\end{stepbox}

\begin{stepbox}
--- Scoring Workflow (shared) --- \hfill \texttt{[08\_scorer.md]}
\begin{enumerate}[leftmargin=1.5em, itemsep=2pt, topsep=4pt]
    \item \textbf{Programmatic Only}: a generated per-round scoring script dynamically imports each scenario's validated \texttt{eval.py} and scores every output; no external APIs, no informal judgment.
    \item \textbf{Structured Failure}: missing outputs, import failures, and truncated generations are scored $0$ with explicit error types, never dropped. 
    \item \textbf{Difficulty Signals}: compute per-task means and cross-model spreads; flag items as \emph{non-discriminative} (spread $\le 0.10$) or \emph{too-easy} (mean $\ge 0.85$ or all pass) for audit. 
    \item \textbf{Output}: per-model score records in \texttt{scores/}, aggregated summaries, and a scoring report.
\end{enumerate}
\end{stepbox}

\subsubsection{Stage 4: Lifecycle Audit}

\begin{stepbox}
--- Lifecycle Audit Workflow (shared) --- \hfill \texttt{[09-12\_*.md]}
\begin{enumerate}[leftmargin=1.5em, itemsep=2pt, topsep=4pt]
    \item \textbf{Failure Analysis}: classify scenarios against the healthy band $[0.25, 0.65]$; separate confound-driven scores from genuine capability signal; isolate high-variance families as capability boundaries.
    \item \textbf{Lifecycle Decision}: assign each scenario one of \textsc{Keep-and-Evolve}, \textsc{Keep-Stable}, \textsc{Reseed}; kept-but-easy scenarios carry explicit escalation instructions (\emph{must-be-harder-than} task IDs) into \texttt{next\_round\_scenario\_plan.json}. 
    \item \textbf{Stop Decision}: continue while valid harder tasks remain producible; termination is triggered only by novelty collapse, evaluator instability, or repeated failure to produce valid harder tasks, never by high scores alone. 
    \item \textbf{Round Summary}: consolidate metrics, audit every executable component (evaluators, runner, scorer) for strict reproducibility, and distill improvement signals for the outer loop. 
\end{enumerate}
\end{stepbox}

\newpage
\subsection{Implementation Details of \sys}

We instantiate the core autonomous agent model ($\mathcal{M}_{\text{agent}}$) of \sys\ using \texttt{Claude-Sonnet-5} to drive the entire workflow end-to-end. The agent's behavior is governed by specific skills and workflows tailored to each of the aforementioned evolutionary stages. Execution begins by downloading the necessary data and parsing the dataset, project scope, and workflow. Within the evolution loop, the proxy models ($\mathcal{M}_1, \mathcal{M}_2$) are accessed via API, with all requests routed through OpenRouter \citep{openrouter2026}. To ensure resource efficiency, the agent's generation is constrained by a 16,000-token output budget and a 600-second timeout per call. A small fraction of API calls yielding errors (e.g., empty responses, timeouts, or budget overflows) are robustly handled, with token counts and scores recorded exactly as returned.

\section{Additional Experimental Details and Results}
\label{app:exp_details}

This appendix provides foundational background and supplementary configurations to offer a more comprehensive understanding of the experimental setups presented in Section~\ref{sec:experiments}. 
It is structured into two primary parts: Appendix~\ref{app:heldout_model_evaluation} details the evaluation protocols and configurations for the LLMs, while Appendix~\ref{app:harness_edits} outlines the experimental setup used to analyze the framework's evolutionary dynamics and rigorously assess the evolved benchmark's quality. 

\subsection{Experimental Setup for Main Result of Held-out Model Evaluation}
\label{app:heldout_model_evaluation}

\textbf{Benchmark Baselines and Different Evolution Rounds. }
To establish a comparative baseline that quantifies the difficulty introduced by \sys, we first evaluate models on the original, unevolved seed datasets (e.g., the original CodeContests and AIME-2024 benchmark) which serve as our baselines. 
To demonstrate how the subsequent evolution process enhances benchmark quality by increasing difficulty and better differentiating model capabilities (results presented in Table~\ref{tab:consolidated_results}), we analyze specific Early and Late milestone rounds. For MB-CodeContests, we select rounds 2 and 10 to represent the Early and Late stages, respectively. For MB-AIME, rounds 1 and 5 are chosen as the corresponding Early and Late stages. 

\textbf{Evolved Benchmarks via Different Evolution Methods. }
To rigorously validate the effectiveness of our approach, \sys, and isolate the performance gains attributed to the evolution method (results presented in Table~\ref{tab:evolution_method}), we compare the \sys-evolved benchmark against datasets generated by several alternative methodologies, all of which are evolved from AIME-2024. 
For external baselines, we generate a comparison dataset using the \texttt{AutoEvoEval}~\citep{wu2025autoevoeval} framework by reimplementing its official specifications. Specifically, we retain its seven question-level operations and rule-based character injection, and randomly sample and perturb scenarios across rounds. Note that we exclude some evolution units that are not related to math or coding. 
Regarding the \texttt{TRACE}~\citep{guo2026towards_trace} methodology, due to its private codebase, we directly report performance metrics from the original paper rather than generating a separate dataset. 
Furthermore, we construct two internal ablation datasets to evaluate specific evolution components of \sys. 
First, we produce a dataset using a \texttt{Naïve Prompting} strategy by entirely removing the outer loop and disabling candidate viability mechanisms (i.e., the model-validation and lifecycle-audit stages), meaning candidates are accepted without intrinsic evaluation. 
Second, we generate a dataset via a \texttt{Fixed Harness} configuration. While this setting also eliminates the outer loop (preventing prompt and configuration edits between rounds), it maintains identical seed items and evaluation logic, thereby isolating the specific impact of harness-level evolution. 

\textbf{Proxy and Held-Out Model Setup.}
To ensure comprehensive coverage of current LLM capabilities and strictly separate the models used for benchmark generation from those used for evaluation (with evaluation results presented in Table~\ref{tab:consolidated_results}), we categorize our models into proxy and held-out sets. 
For the \textbf{proxy models}, which drive the in-loop evolution to ensure the generated candidates successfully meet our quality and viability expectations, we utilize \texttt{Claude-Sonnet-5} as an Adaptive Reasoning model (exhibiting dynamic reasoning and tool-use self-correction) and \texttt{GPT-5.4} as a Standard Instruct (Non-Reasoning) model. 
Conversely, for the \textbf{held-out models}, which are evaluated on the resulting benchmarks to verify that the evolved problems effectively enhance discriminative power across unseen architectures, we employ two distinct behavioral classes: Standard Instruct models operating under standard inference budgets (e.g., \texttt{DeepSeek-V3.1}, \texttt{Kimi-K2.5}, \texttt{Gemini-2.5-Flash}), and Test-Time Scaling (Reasoning) models (e.g., \texttt{DeepSeek-V3.1}, \texttt{Kimi-K2.5}, \texttt{Gemini-2.5-Flash}, \texttt{DeepSeek-R1-Distill-Qwen-7B/32B},), which are specialized for long-horizon thinking, self-reflection, and chain-of-thought scaling. 
As introduced in Section~\ref{subsec:main_results}, all held-out model evaluations employ standard generation constraints. 
Notably, the two Qwen reasoning models are evaluated across two strictly budgeted phases: a reasoning phase capped at 96,000 tokens, followed by an answer-extraction phase capped at 32,000 tokens, with the final answer derived exclusively from the second phase.

\subsection{Experimental Setup for Evolutionary Dynamics and Quality Analysis}
\label{app:harness_edits}

This section analyzes the internal dynamics of our evolutionary framework and the intrinsic quality of the resulting benchmarks. We begin by examining the meta-harness's outer-loop adaptations in Appendix~\ref{app:harness_edits}. Next, Appendices~\ref{app:harness_evo_dynamics} and \ref{app:breakdown_analysis} explore task generation qualitatively and quantitatively. To ensure data integrity, Appendix~\ref{app:quality_assessment} rigorously assesses benchmark quality, focusing on reasonableness, competency, and robustness. Finally, Appendix~\ref{app:appendix_case_study} grounds these analyses with detailed case studies, illustrating the trajectory from seed problems to fully evolved tasks. 



\subsubsection{Harness-Level Evolutionary Dynamics Analysis}
\label{app:harness_evo_dynamics}

While Section~\ref{subsec:harness_evolution} analyzes how benchmark harness edits drive domain-specific adaptations, this appendix provides the extended details that underpin the benchmark harness-edit audit (Table~\ref{tab:meta_change_stats}). 
Specifically, our audit draws on three archived artifacts per candidate, namely the machine-readable proposals, per-round leaderboards, and apply reports, to categorize harness modifications into four functional groups. These groups comprise \textit{measurement} updates for evaluators and scorers, \textit{generation} modifications for task evolution and seed selectors, \textit{execution} adjustments governing runners and configurations, and \textit{telemetry} refinements for analyzers and early-stopping conditions. 
Beyond the domain-adaptive trends discussed in Section~\ref{subsec:harness_evolution}, this fine-grained granular breakdown reveals that a steady minority of edits is actively dedicated to maintaining the instrument itself. Crucially, these maintenance updates introduce execution-reliability fixes like robust timeout handling and retry safety mechanisms, alongside telemetry rules designed to filter out anomalous or confounded scores from lifecycle decisions. Ultimately, these adjustments highlight an organic synergy within the dual-loop workflow. 
By continuously reinforcing the evaluation environment and filtering unreliable feedback, the outer harness-editing loop ensures that the inner loop can explore increasingly complex problem spaces, thereby driving a stable and continuous evolution. 

\subsubsection{Qualitative Analysis on Task Evolution}
\label{app:qualitative_analysis_and_case_study}

\begin{figure}[htbp]
    \centering
    \begin{minipage}{0.96\linewidth}
        \centering
        \includegraphics[width=\linewidth]{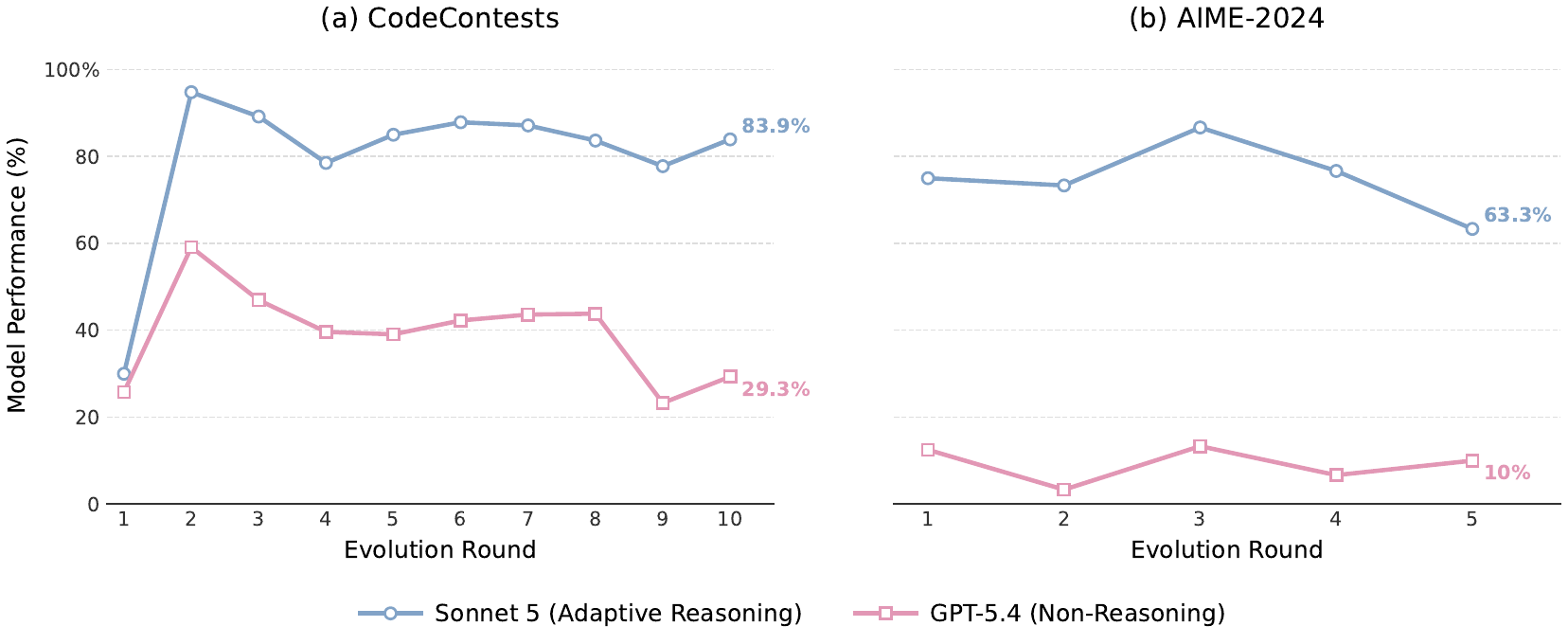}
    \end{minipage}
    \vspace{-0.5em}
    \caption{Round-by-round model validation scores along the evolution of \sys. }
    \label{fig:qual_line_charts}
\end{figure}

While the statistical breakdowns in Section~\ref{subsec:harness_evolution} identify which harness components evolved, the main purpose of this section is to examine the model validation feedback to understand how the system simultaneously escalates task difficulty and preserves model discriminability. Specifically, we aim to demonstrate that the meta-harness does not merely apply random mutations, but dynamically reacts to the proxy models' behaviors by adjusting constraints, fixing formats, and pushing capability ceilings round by round. To capture this interactive process, Figure~\ref{fig:qual_line_charts} tracks the longitudinal in-loop validation scores for our proxy models (\texttt{Claude-Sonnet-5} and \texttt{GPT-5.4}), illustrating two distinct evolutionary trajectories for CodeContests and AIME-2024.

\noindent\textbf{Coding Problems. } 
The 10-round CodeContests evolution (Figure~\ref{fig:qual_line_charts}a) reveals a dynamic shift from evaluator correction to difficulty escalation, unfolding in four distinct phases: 
\begin{itemize}[leftmargin=1.5em, topsep=2pt, itemsep=2pt]
    \item \textit{Round 1 (Identifying Measurement Defects):} The evolution initially records artificially low performance, with \texttt{Claude-Sonnet-5} and \texttt{GPT-5.4} scoring 0.300 and 0.258. The failure analyzer reveals that out of 180 model-task pairs, 121 fail due to output formatting (e.g., answers wrapped in explanatory text) rather than flawed reasoning, masking the models' true capabilities. 
    
    \item \textit{Round 2 (Evaluator Correction):} Recognizing that the defect lies in the evaluator rather than the task, the meta-harness strictly corrects the measurement logic by mandating explicit answer markers. Without altering actual task difficulty, scores surge to 0.948 and 0.591, proving that the seed tasks are actually too easy once superficial formatting confounds are removed.
    
    \item \textit{Rounds 3--4 (Complexity Escalation):} Having secured a robust evaluation mechanism, the framework shifts to escalating task complexity, successfully driving performance down to 0.785 and 0.396. This targeted escalation establishes a clean capability split across three specific scenarios, which \texttt{Claude-Sonnet-5} solves perfectly while \texttt{GPT-5.4} fails entirely.
    
    \item \textit{Rounds 5--10 (Refinement and Final Separation):} 
    Although Rounds 5--7 experience a temporary stall, the system learns from ineffective mutations (e.g., asking a model to certify logic it has already mastered, which inadvertently simplifies the task) and retires scenarios that fail to increase difficulty after two consecutive attempts. Leveraging these insights, Rounds 8--9 intensify the remaining tasks. Ultimately, by Round 10, the performance gap between the two models widens significantly to 0.546 (0.839 vs. 0.293), compared to a mere 0.042 margin in Round 1, achieving the dual objective of escalating difficulty while preserving discriminability. 
\end{itemize}

\noindent\textbf{Mathematical Problems. } 
Unlike CodeContests, the 5-round AIME trajectory (Figure~\ref{fig:qual_line_charts}b) bypasses initial measurement defects to directly target mathematical reasoning, unfolding as follows: 
\begin{itemize}[leftmargin=1.5em, topsep=2pt, itemsep=2pt]
    \item \textit{Round 1 (Immediate Capability Separation):} The initial round separates the models (0.750 vs.\ 0.125). Unlike CodeContests, \texttt{GPT-5.4}'s failures stem from incorrect numeric derivations rather than format rejections, allowing the framework to proceed to difficulty escalation. 
    
    \item \textit{Rounds 2--3 (Isolating Reasoning Gaps):} Round 2 (0.733 vs.\ 0.033) highlights a specific problem cluster where both models score 0.0 due to a shared misconception regarding an evolved combinatorial family (rigorously verified against ground truth). In Round 3 (0.867 vs.\ 0.133), a performance divergence becomes the informative event: \texttt{Claude-Sonnet-5} overcomes the misconception, whereas \texttt{GPT-5.4} repeats the identical error. This effectively localizes the performance gap to a specific reasoning step rather than general difficulty.

    \item \textit{Round 4 (Handling Confounds):} Recording the peak performance divergence of the study (0.767 vs.\ 0.067), this round encounters infrastructure-level confounds (e.g., programmatic execution timeouts). Instead of absorbing these anomalies into the final scores, the evaluation loop correctly identifies the issue and pauses difficulty escalation on the affected scenarios. 
    
    \item \textit{Round 5 (Correcting Instruction Drift \& Final Bounds):} 
    Delivering the first fully clean run since Round 3 (0.633 vs.\ 0.100), the system diagnoses an apparent capability ceiling as ``instruction drift'' (where a previously proven step-by-step reasoning constraint had been silently dropped). By restoring this constraint, the framework avoids mistaking prompt saturation for inherent task difficulty. Across the entire trajectory, \texttt{GPT-5.4} never exceeds a score of 0.133, while \texttt{Claude-Sonnet-5} consistently remains above 0.633.
\end{itemize}


Together, these round-by-round case studies concretely illustrate the dual-loop synergy introduced in Appendix~\ref{app:harness_evo_dynamics}. 
By adaptively distinguishing between basic measurement errors (e.g., evaluator flaws, instruction drift) and actual mathematical or algorithmic reasoning limits, the outer harness-editing loop continuously stabilizes the evaluation environment. This targeted filtering of unreliable feedback ensures that the inner loop can safely escalate task complexity without being derailed by superficial confounds, ultimately generating capability gaps that accurately reflect the models' true performance boundaries.

\subsubsection{Quantitative Analysis of Evolved Benchmark Characteristics}
\label{app:breakdown_analysis}

\begin{figure}[htbp]
    \centering
    \includegraphics[width=\columnwidth]{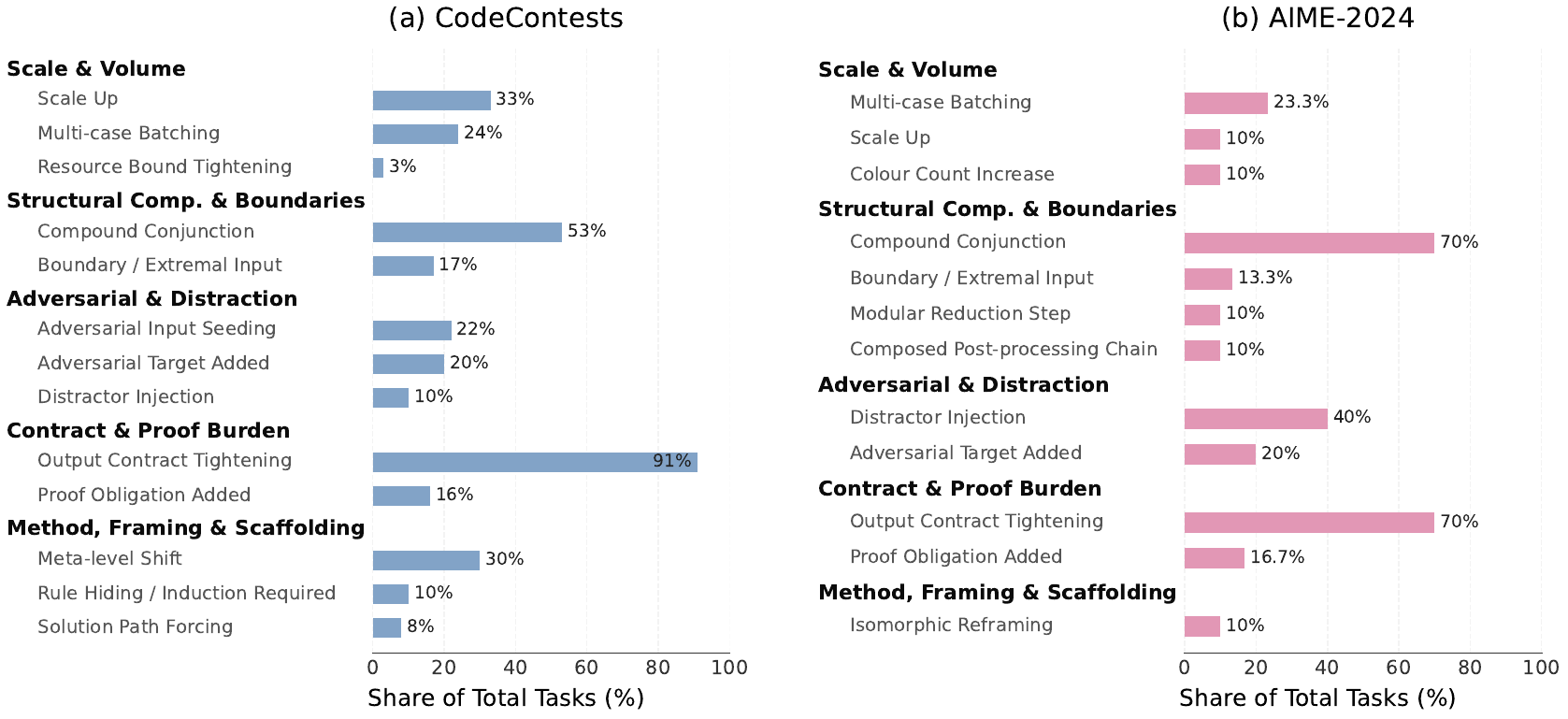}
    \caption{A detailed quantitative analysis of the distribution of difficulty levers. }
    \label{fig:difficulty_levers_dist}
\end{figure}

\begin{figure}[htbp]
    \centering
    \includegraphics[width=\columnwidth]{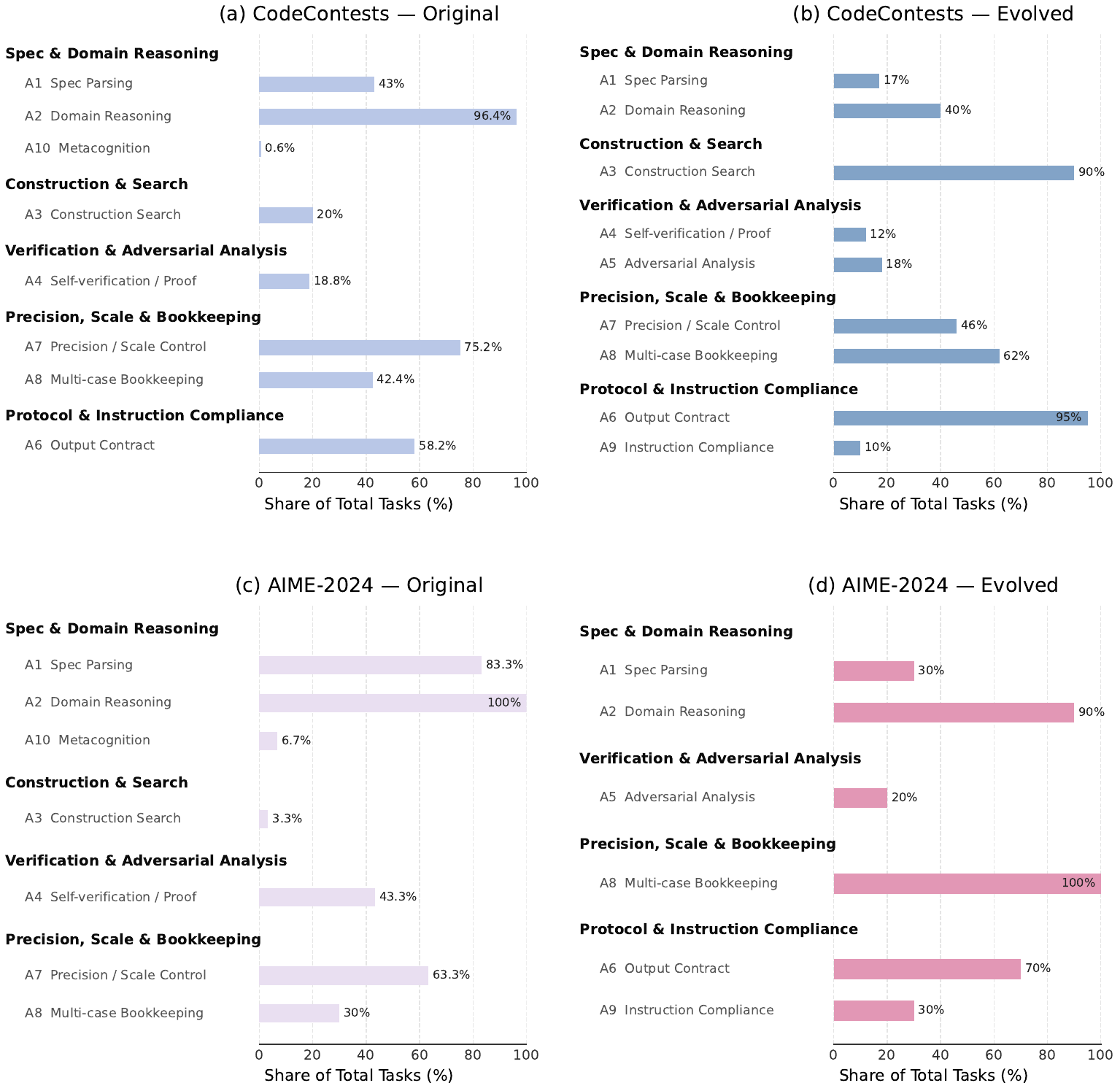}
    \caption{A detailed quantitative analysis of the distribution of the capability distribution. }
    \label{fig:capabilities_dist}
\end{figure}

\begin{figure}[htbp]
    \centering
    \includegraphics[width=\columnwidth]{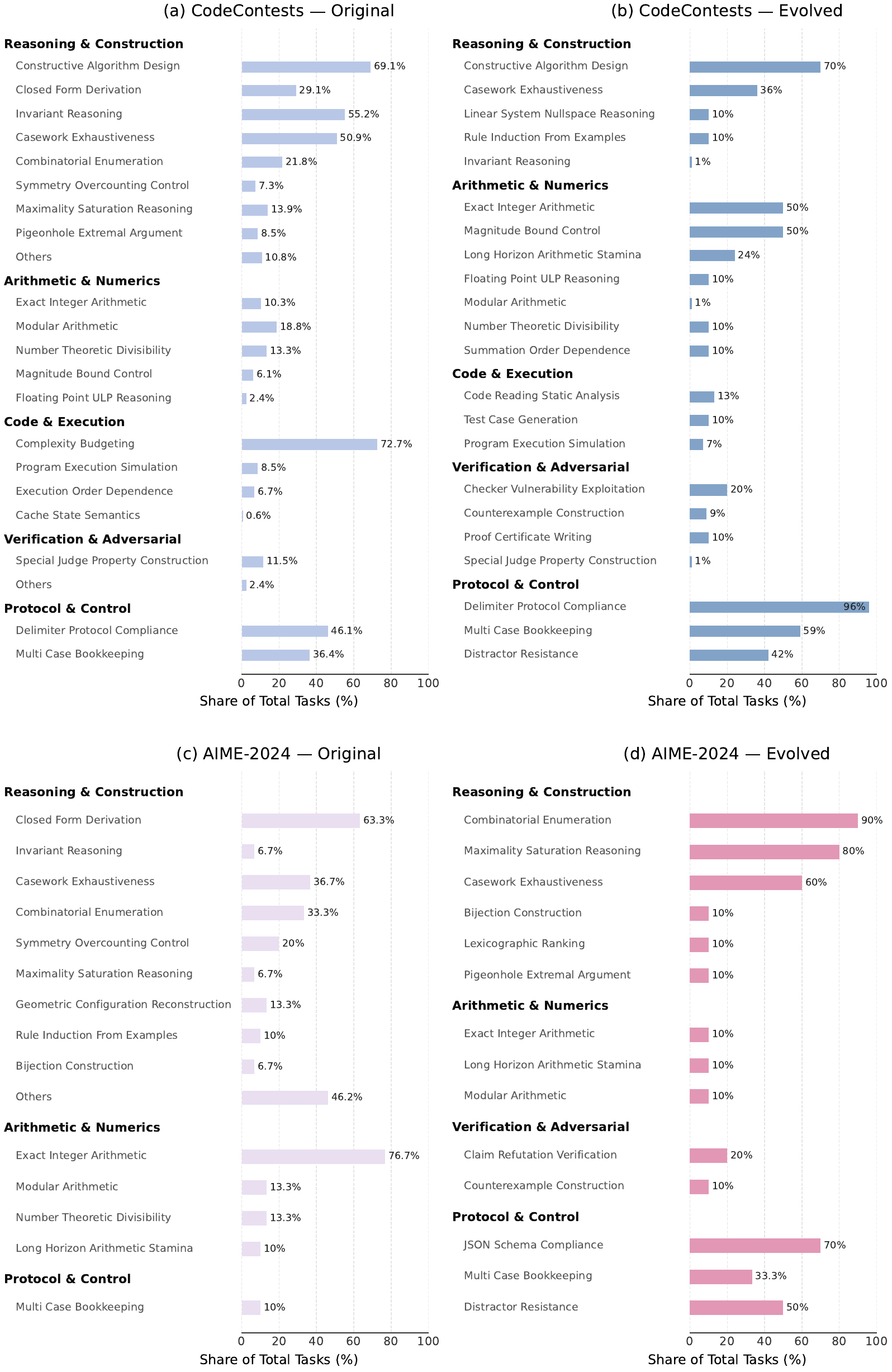}
    \caption{A detailed quantitative analysis of the distribution of the problem-solving technique. }
    \label{fig:solving_technique_dist}
\end{figure}

Building on Section~\ref{subsec:harness_evolution} regarding which dimensions of the benchmark's capabilities are evolved, this section provides a fine-grained quantitative analysis of the resulting tasks. Specifically, we evaluate the benchmarks along three complementary dimensions: \textit{Difficulty Levers}, \textit{Capabilities}, and \textit{Solving Techniques}. 
These taxonomies characterize the evolution at different angles, 
capturing the specific evolutionary mechanisms chosen to escalate difficulty (Figure~\ref{fig:difficulty_levers_dist}), and tracking the distributional shifts from original to evolved tasks in both evaluated capabilities (Figure~\ref{fig:capabilities_dist}) and underlying solving techniques (Figure~\ref{fig:solving_technique_dist}). 
Difficulty levers denote the operational methods used to mutate seed problems, whereas the other two dimensions compare the original and evolved benchmarks. 

\noindent\textbf{Distribution of Difficulty Levers. } 
Figure~\ref{fig:difficulty_levers_dist} illustrates how the meta-harness applies domain-specific mechanisms to escalate task difficulty. To actively drive complexity, the framework relies on structural levers such as ``Output Contract Tightening,'' ``Compound Conjunction,'' and ``Distractor Injection.'' The AIME-2024 trajectory aggressively utilizes these active mutations: mathematical difficulty is primarily driven by ``Output Contract Tightening'' (70\%) and ``Compound Conjunction'' (70\%), alongside adversarial reframing like ``Distractor Injection'' (40\%). Similarly, in CodeContests, ``Output Contract Tightening'' (91\%) and ``Compound Conjunction'' (53\%) serve as the primary structural drivers, alongside ``Scale Up'' (33\%) and ``Meta Level Shift'' (30\%). However, a domain disparity emerges: mathematical evolution relies more heavily on structural and adversarial reframing, whereas coding evolution spreads its difficulty across a broader range of mechanisms, including ``Multi-case Batching'' (24\%) and ``Adversarial Input Seeding'' (22\%). This indicates that both domains introduce substantive changes beyond mere instance regeneration, while relying on distinct mechanisms to escalate complexity.

\noindent\textbf{Shift in Evaluated Capabilities. } 
Figure~\ref{fig:capabilities_dist} illustrates the evolutionary shift in the breadth of capabilities required by the benchmarks. To escalate cognitive burden, the framework actively enforces exhaustive state-tracking and strict formatting constraints. In CodeContests, the system pivots toward rigorous algorithmic implementation, driving ``Construction Search'' to surge from 20\% to 90\% and ``Output Contract'' compliance to reach 95\%; consequently, the original reliance on pure ``Domain Reasoning'' (96.4\%) drastically drops. Similarly, AIME-2024 aggressively amplifies multi-step execution demands: ``Multi-case Bookkeeping'' skyrockets from 30\% to 100\% and ``Output Contract'' emerges at 70\%. While AIME maintains its foundational ``Domain Reasoning'' (90\%), it decisively sheds basic ``Spec Parsing'' (dropping from 83.3\% to 30\%). Across both domains, this dynamic demonstrates that successful benchmark evolution builds toward exhaustive search, complex execution, and strict protocol compliance, thereby outgrowing the need to test basic conceptual parsing. 

\noindent\textbf{Evolution of Solving Techniques. } 
Figure~\ref{fig:solving_technique_dist} details the shift in the underlying techniques required to crack individual tasks. To actively push models to their limits, the framework enforces rigorous combinatorial searches and survival against adversarial constraints. In CodeContests, evolution creates a massive demand for ``Delimiter Protocol Compliance'' (96\%) and explicit ``Magnitude Bound Control'' (50\%), completely replacing the original dependency on standard ``Complexity Budgeting'' (which vanishes from 72.7\%). Similarly, evolved mathematical tasks in AIME-2024 heavily enforce grueling logical searches dominated by ``Combinatorial Enumeration'' (90\%) and ``Maximality Saturation Reasoning'' (80\%), alongside strict ``JSON Schema Compliance'' (70\%). To make room for these advanced demands, the trajectory decisively abandons standard textbook approaches: initial requirements like ``Closed Form Derivation'' (63.3\%) disappear entirely, while ``Exact Integer Arithmetic'' plummets from 76.7\% to just 10\%. Ultimately, by forcing models to master complex casework, distractor resistance, and unyielding protocol adherence, the framework systematically evolves both domains away from straightforward heuristic problem-solving. 

Overall, the distributions confirm that \sys\ actively drives domain-specific evolution to increase complexity. In CodeContests, the framework enforces a constraint-based evolution, maximizing difficulty through strict protocol requirements, compound constraints, and meta-level reframing. In contrast, AIME-2024 undergoes active structural evolution, where the system injects distractors and domain-specific structural mutations (e.g., ``Modular Reduction Step'') and demands heavy combinatorial casework to build upon the core mathematical foundation.
Despite these differences, both domains show that \sys\ systematically enriches task composition rather than simply lengthening reasoning, adaptively tailoring its mechanisms to each domain's natural demands.

\newpage

\subsubsection{Quality Assessment of the Evolved Benchmarks}
\label{app:quality_assessment}
\begin{figure}[htbp]
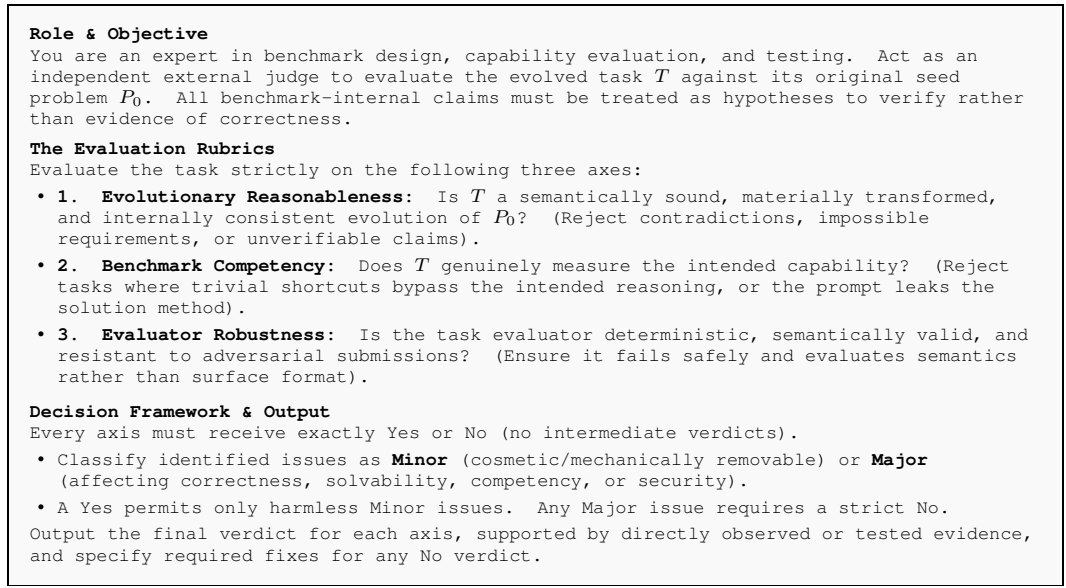

\centering
\begin{promptbox}
\textbf{Role \& Objective} \\
You are an expert in benchmark design, capability evaluation, and testing. 
Act as an independent external judge to evaluate the evolved task $T$ against its original seed problem $P_0$. All benchmark-internal claims must be treated as hypotheses to verify rather than evidence of correctness.

\smallskip
\textbf{The Evaluation Rubrics} \\
Evaluate the task strictly on the following three axes:
\begin{itemize}[leftmargin=1.5em, itemsep=0pt, topsep=2pt]
    \item \textbf{1. Evolutionary Reasonableness:} Is $T$ a semantically sound, materially transformed, and internally consistent evolution of $P_0$? (Reject contradictions, impossible requirements, or unverifiable claims).
    \item \textbf{2. Benchmark Competency:} Does $T$ genuinely measure the intended capability? (Reject tasks where trivial shortcuts bypass the intended reasoning, or the prompt leaks the solution method).
    \item \textbf{3. Evaluator Robustness:} Is the task evaluator deterministic, semantically valid, and resistant to adversarial submissions? (Ensure it fails safely and evaluates semantics rather than surface format).
\end{itemize}

\smallskip
\textbf{Decision Framework \& Output} \\
Every axis must receive exactly \texttt{Yes} or \texttt{No} (no intermediate verdicts). 
\begin{itemize}[leftmargin=1.5em, itemsep=0pt, topsep=2pt]
    \item Classify identified issues as \textbf{Minor} (cosmetic/mechanically removable) or \textbf{Major} (affecting correctness, solvability, competency, or security).
    \item A \texttt{Yes} permits only harmless Minor issues. Any Major issue requires a strict \texttt{No}.
\end{itemize}
Output the final verdict for each axis, supported by directly observed or tested evidence, and specify required fixes for any \texttt{No} verdict.
\end{promptbox}
\caption{The simplified version system prompt for the quality evaluation using LLM-as-a-judge. }
\label{fig:judge_prompt}
\end{figure}

\begin{figure}[h]
    \centering
    \vspace{-2em}
    \begin{minipage}{\linewidth}
        \centering
        \includegraphics[width=0.96\linewidth]{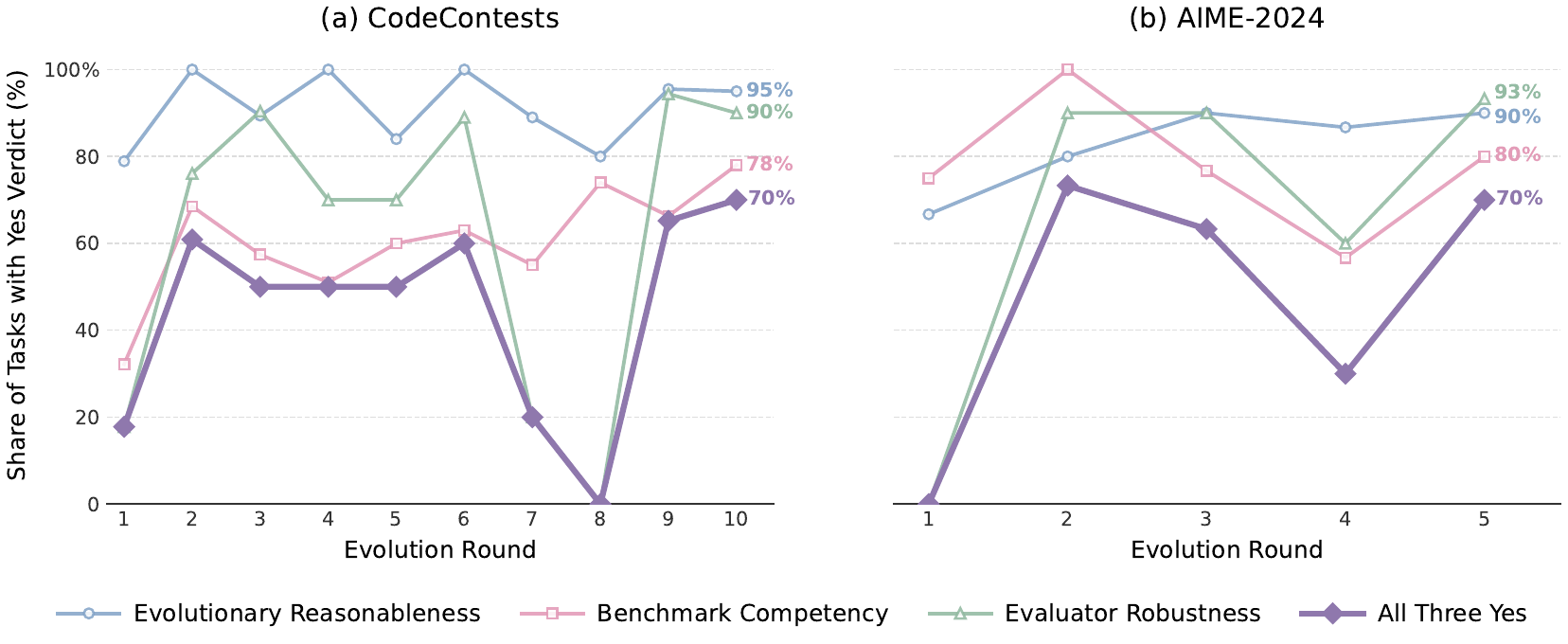}
    \end{minipage}
    \vspace{-0.5em}
    \caption{Quality assessment of evolved benchmarks across iterative rounds. }
    \label{fig:app_line_charts}
\end{figure}

Building on Section~\ref{subsec:harness_evolution}, this section evaluates task quality across every evolutionary round rather than focusing solely on late-round outcomes. 
An isolated LLM judge (\texttt{Claude-Opus-5}) assesses each evolved task against its seed across Reasonableness, Competency, and Robustness (Table~\ref{tab:validity_audit}). Figure~\ref{fig:judge_prompt} provides the simplified version system prompt, 
detailing the essential evaluation criteria while omitting operational overhead like I/O formatting.
Figure~\ref{fig:app_line_charts} reveals benchmark evolution as an aggressively exploratory, non-monotonic process. While ``Evolutionary Reasonableness'' remains consistently high, pushing difficulty boundaries causes severe fluctuations in stricter metrics. Sudden metric drops, such as the collapse in ``Evaluator Robustness'' during CodeContests Round 8 and AIME Round 4, do not indicate systemic failure; rather, they highlight phases where extreme structural mutations temporarily break solver reliability. Crucially, the system exhibits strong self-correction. Following these exploratory crashes, both trajectories recover entirely, converging to robust final metrics (e.g., $>90\%$ Evaluator Robustness) and a strict 70\% overall acceptance rate (``All Three Yes''). This volatile trajectory proves the framework actively risks temporary quality degradation to discover deeper complexities before successfully stabilizing.

\subsubsection{Detailed Case Studies: From Seed Tasks to Evolved Tasks}
\label{app:appendix_case_study}
While Table~\ref{tab:merged_evolution_case_study} summarizes the overall evolution, Table~\ref{tab:merged_evolution_case_study_appendix} provides a comprehensive, round-by-round breakdown for both the MB-CodeContests and MB-AIME scenarios. It explicitly tracks the transformation of the Task Formulation from the initial seed to the deepened stages. Additionally, the table unpacks the progressive expansion of required Capabilities and Solving Techniques to illustrate how new cognitive demands are introduced. Finally, it enumerates the specific Applied Difficulty Levers used at each round to systematically drive this multi-dimensional evolution.

\begin{table*}[htbp]
\centering
\small
\caption{\sys\ Evolution Case Study: From Initial Seed to Deepening}
\vspace{-0.8em}
\label{tab:merged_evolution_case_study_appendix}

\begin{tabularx}{\textwidth}{>{\RaggedRight\hsize=1.0\hsize}X >{\RaggedRight\hsize=1.0\hsize}X >{\RaggedRight\hsize=1.0\hsize}X}
\multicolumn{3}{c}{Part I: Case Study of MB-CodeContests Evolution} \\
\toprule
\textbf{Seed} (Codeforces 1582D) & \textbf{Round 5} (Evolved Emergence) & \textbf{Round 9} (Evolved Deepening) \\
\midrule

\rowcolor{gray!20}
\multicolumn{3}{l}{\textit{Task Formulation}} \\
\addlinespace[2pt]
\textbf{Constructive Math}: Given an array $a$ ($a_i \neq 0$), construct $b$ such that $\sum a_i b_i = 0$ and $\sum |b_i| \le 10^9$. & 
\textbf{Single Gap Audit}: Exploit a planted gap (\texttt{range(n-1)}) to trick a Python checker with an invalid $(a, b)$ pair. &
\textbf{Compound Audit}: Trace complex logic to exploit a vulnerability gated by 3 conditions ($n > 1000$, duplicate $a$, odd $n$). \\
\addlinespace[4pt]

\midrule
\rowcolor{gray!20}
\multicolumn{3}{l}{\textit{Capabilities and Solving Techniques}} \\
\addlinespace[2pt]
\textbf{Capabilities:} \newline Construction search, \par Domain reasoning,  \par Precision scale control \par\smallskip
\textbf{Solving Techniques:} \newline Constructive algorithm design, Exact integer arithmetic, \par Magnitude bound control 
&
\textbf{Capabilities:} \newline Adversarial analysis, Construction search, Output contract \par\smallskip
\textbf{Solving Techniques:} \newline Code reading static analysis, Checker vulnerability exploitation, Counterexample construction, Program execution simulation, Protocol compliance 
&
\textbf{Capabilities:} \newline Adversarial analysis, Construction search, Precision scale control, Output contract \par\smallskip
\textbf{Solving Techniques:} \newline Code reading static analysis, Checker vulnerability exploitation, Counterexample construction, Long horizon arithmetic stamina, Casework exhaustiveness, Protocol compliance \\
\addlinespace[4pt]

\midrule
\rowcolor{gray!20}
\multicolumn{3}{l}{\textit{Applied Difficulty Levers}} \\
\addlinespace[2pt]
Baseline
&
Meta level shift, Adversarial target added, Compound conjunction, Output contract tightening
&
Meta level shift, Adversarial target added, Output contract tightening, Compound conjunction, Scale up, Scaffolding removal \\
\bottomrule
\end{tabularx}

\vspace{0.5em} 

\begin{tabularx}{\textwidth}{>{\RaggedRight\hsize=1.0\hsize}X >{\RaggedRight\hsize=1.0\hsize}X >{\RaggedRight\hsize=1.0\hsize}X}
\multicolumn{3}{c}{Part II: Case Study of MB-AIME Evolution} \\
\toprule
\textbf{Seed} (AIME 2024-II-9) & \textbf{Round 1} (Evolved Emergence) & \textbf{Round 5} (Evolved Deepening) \\
\midrule

\rowcolor{gray!20}
\multicolumn{3}{l}{\textit{Task Formulation}} \\
\addlinespace[2pt]
\textbf{Combinatorial Counting}: Count placements on a $5 \times 5$ grid where no further chip can be added (\emph{answer}: $902$). &
\textbf{Pre-Blocked Obstacle}: Count maximal placements on a $4 \times 4$ board with 2 permanently blocked cells (\emph{answer}: $150$). &
\textbf{Post-Hoc Filter}: Count maximal placements on a $3 \times 5$ board, then filter out those hitting 4 flagged cells (\emph{answer}: $14$). \\
\addlinespace[4pt]

\midrule
\rowcolor{gray!20}
\multicolumn{3}{l}{\textit{Capabilities and Solving Techniques}} \\
\addlinespace[2pt]
\textbf{Capabilities:} \newline Domain reasoning, Multi-case bookkeeping \par\smallskip
\textbf{Solving Techniques:} \newline Maximality saturation reasoning, Combinatorial enumeration, Casework exhaustiveness, \par Symmetry overcounting control 
&
\textbf{Capabilities:} \newline Domain reasoning, Multi-case bookkeeping, Spec parsing \par\smallskip
\textbf{Solving Techniques:} \newline Combinatorial enumeration, Maximality saturation reasoning, Casework exhaustiveness 
&
\textbf{Capabilities:} \newline Domain reasoning, \par Multi-case bookkeeping, \par Instruction compliance \par\smallskip
\textbf{Solving Techniques:} \newline Combinatorial enumeration, Maximality saturation reasoning, Distractor resistance, \par Casework exhaustiveness \\
\addlinespace[4pt]

\midrule
\rowcolor{gray!20}
\multicolumn{3}{l}{\textit{Applied Difficulty Levers}} \\
\addlinespace[2pt]
Baseline
&
Compound conjunction
&
Distractor injection, \par
Solution path forcing \\
\bottomrule
\end{tabularx}
\end{table*}

\raggedbottom

\end{document}

%% file: math_commands.tex
\usepackage{amsmath,amsfonts,bm}

\def\eqref#1{equation~\ref{#1}}

\def\1{\bm{1}}

\DeclareMathAlphabet{\mathsfit}{\encodingdefault}{\sfdefault}{m}{sl}
\SetMathAlphabet{\mathsfit}{bold}{\encodingdefault}{\sfdefault}{bx}{n}

